\documentclass[10pt,twocolumn,letterpaper]{article}

\usepackage{iccv}
\usepackage{times}
\usepackage{epsfig}
\usepackage{graphicx}
\usepackage{amsmath}
\usepackage{amssymb}

\usepackage{xspace}
\usepackage{mathrsfs}
\usepackage{amsmath, amssymb, amsfonts, bm}
\usepackage[flushleft]{threeparttable}
\usepackage{tablefootnote, subcaption}
\usepackage{multirow}
\usepackage{hhline}
\usepackage{graphicx}
\usepackage{hanging}
\usepackage{color}

\usepackage[table]{xcolor}
\usepackage{float}

\usepackage{wrapfig}
\usepackage{balance}
\usepackage{makecell, booktabs}
\usepackage{hhline}
\usepackage{ifthen}
\usepackage{caption}
\usepackage{graphics} % for pdf, bitmapped graphics files
\usepackage{epsfig} % for postscript graphics files
\usepackage{mathptmx} % assumes new font selection scheme installed
\usepackage{times} % assumes new font selection scheme installed

\usepackage{enumitem}

\usepackage{pifont}  % check and cross
\usepackage{txfonts}

\usepackage[accsupp]{axessibility}
\newcommand{\methName}{WDiscOOD\xspace}

\newcommand{\feat}{\bm{x}}
\newcommand{\Dim}{D}
\newcommand{\cls}{c}
\newcommand{\clsNum}{C}
\newcommand{\trainNum}{N}

\newcommand{\score}[1][]{\ifthenelse{\equal {#1} {}}{s(\cdot)}{s({#1})}\xspace}
\newcommand{\gSpace}{Whitened Discriminative Subspace\xspace}
\newcommand{\hSpace}{Whitened Discriminative Residual Subspace\xspace}
\newcommand{\gSpaceS}{WD\xspace}
\newcommand{\hSpaceS}{WDR\xspace}
\newcommand{\WLDA}{Whitened Linear Discriminant Analysis\xspace}
\newcommand{\WLDAS}{WLDA\xspace}
\newcommand{\discriminant}{\bm{w}}
\newcommand{\Sw}{\bm{S}_w}
\newcommand{\Sb}{\bm{S}_b}
\newcommand{\cen}{\bm{\mu}}
\newcommand{\encoder}[1][]{\ifthenelse{\equal {#1} {}}{\phi(\cdot)}{\phi({#1})}\xspace}

\newcommand{\eigVecMat}{\bm{V}}
\newcommand{\eigValMat}{\bm{\Lambda}}
\newcommand{\discNum}{N_{D}}
\newcommand{\gMat}{\bm{W}}
\newcommand{\hWeight}{\alpha}

\newcommand{\eye}{\bm{I}}
\newcommand{\Real}{\mathbb{R}}

\newcommand{\norm}[1]{\|{#1}\|}

\newcommand{\cmark}{\ding{51}}%
\newcommand{\xmark}{\ding{55}}%

\makeatletter
\@namedef{ver@everyshi.sty}{}
\makeatother
\usepackage{balance}
\usepackage{enumitem}
\usepackage{tikz}
\usetikzlibrary{calc,arrows,decorations.markings}

\usepackage[pagebackref=true,breaklinks=true,letterpaper=true,colorlinks,bookmarks=false]{hyperref}

\iccvfinalcopy % *** Uncomment this line for the final submission

\def\iccvPaperID{12110} % *** Enter the ICCV Paper ID here

\ificcvfinal\fi

\begin{document}

%%%%%%%%% TITLE
\title{\methName: Out-of-Distribution Detection via Whitened Linear Discriminant Analysis}

\author{
Yiye~Chen
\quad Yunzhi~Lin
\quad Ruinian~Xu 
\quad Patricio~A.~Vela \\
{Georgia Institute of Technology} \\
{\tt\small \{yychen2019, yunzhi.lin, rnx94, pvela\}@gatech.edu}
}

%\author{
%Ze~Liu\textsuperscript{\dag\thanks{Equal contribution. \textsuperscript{\dag}Interns at MSRA. \textsuperscript{\ddag}Contact person.}}
%\quad Yutong~Lin\textsuperscript{\dag*} \quad Yue~Cao\textsuperscript{*}
%\quad Han~Hu\textsuperscript{*\ddag}
%\quad Yixuan~Wei\textsuperscript{\dag} \\
%\quad Zheng~Zhang
%\quad Stephen~Lin
%\quad Baining~Guo\\
%{Microsoft Research Asia} \\
%\small{\texttt{\{v-zeliu1,v-yutlin,yuecao,hanhu,v-yixwe,zhez,stevelin,bainguo\}@microsoft.com}}
%}

\maketitle
\ificcvfinal\thispagestyle{empty}\fi

%%%%%%%%% ABSTRACT

% Just say we explore feature-space method. Don't justify it over logit/prob space as it is already widely explored. Can mention this in introduction.

% say we discover that whitened discriminative null space is very helpful. Then say "which is applicable for both classifiers and stand-alone visual encoders.

\begin{abstract}
   Deep neural networks are susceptible to generating overconfident yet erroneous predictions when presented with data beyond known concepts.
This challenge underscores the importance of detecting out-of-distribution (OOD) samples in the open world.
%	\red{Previous work has shown great success in combining both class-specific and class-agnostic information, albeit with the need to query classifier head response. 
%	This requirement is not feasible for fast-developing stand-alone pretrained visual encoders.}
  	In this work, we propose a novel feature-space OOD detection score 
    based on class-specific and class-agnostic information.
  	Specifically, the approach utilizes \WLDA to project features into two
    subspaces - the discriminative and residual subspaces - for which the
    in-distribution (ID) classes are maximally separated and closely
    clustered, respectively. 
  	The OOD score is then determined by combining the deviation from the input data to the ID pattern in both subspaces.
   The efficacy of our method, named \methName, is verified on the
   large-scale ImageNet-1k benchmark, with six OOD datasets that cover a
   variety of distribution shifts.
   \methName demonstrates superior performance on deep classifiers with diverse backbone architectures, including CNN and vision transformer.
   Furthermore, we also show that \methName more effectively detects novel
   concepts in representation spaces trained with contrastive objectives,
   including supervised contrastive loss and multi-modality contrastive
   loss.
   
\end{abstract}

%%%%%%%%% BODY TEXT

\begin{figure}[t!]
   \vspace{-10pt}
  % \vspace*{-0.0in}
  \centering
  \scalebox{0.9}{
    \begin{tikzpicture}
     \node[anchor=north west](graph) at (0in,0in)
      {{\includegraphics[width=0.5\textwidth,trim={0.0in
      0.0in 0.0in 0.0in},clip=true]{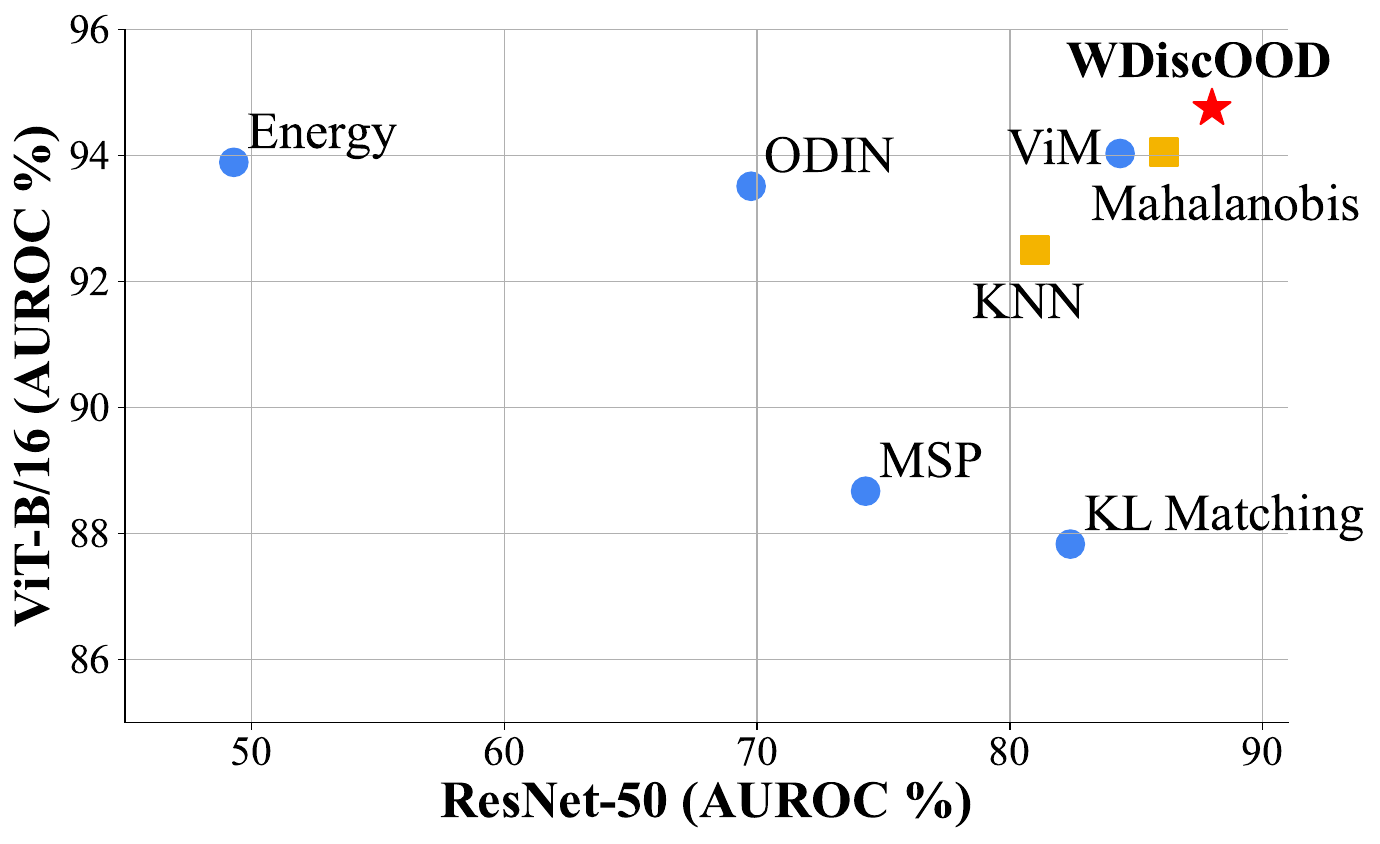}}};
    % space annotation
%    \node[anchor=south] at ($ (graph.west) + (2.2, 1.3)$)  {\small $\enModel_\mParam(\hat{\btau{}}_{i-1})$};
%    \node[anchor=south] at ($ (graph.west) + (4.2, 1.3)$)  {\small $\enModel_\mParam(\hat{\btau{}}_{i})$};

    \end{tikzpicture}
  }
  \vspace{-12pt}
  \caption{\small
  \textbf{Performance on two OOD detection benchmarks.}
  We propose a novel OOD detection method based on \textit{\WLDA}, wihch is highlighted with red star. 
  It outperforms baselines with both ResNet-50 ($x$-axis) and ViT-B/16 ($y$-axis) model on ImageNet-1k benchmark.
  Blue dots and yellow square denote classifier-based and feature-based methods, respectively.
  The AUROCs are averaged over six OOD datasets, for which the detailed results are tabulated in Tab. \ref{tab:clfResults}.
  }
 \label{fig:teaser}
 \vspace{-10pt}
\end{figure}

\section{Introduction}
% Will create a teaser figure. Intro + abstract + teaser figure should go to ~1.75 page
% Structure: 1. OOD is important - close world v.s. open world assumption 
%			2. Best baseline comes from combining both class-specific and class-agnosit information, but requires class head
%			3. We propose a new method that based solely in the feature-space.

% The problem is important
Deep learning models are typically designed with a \textit{closed-world}
assumption \cite{surveyGenOOD}, where test data is assumed to be drawn from the same distribution as the training data. 
Without any built-in mechanism to distinguish novel concepts, deep neural networks are prone to generate incorrect answers with high confidence when presented with \textit{out-of-distribution (OOD)} data.
Such a misleading decision could result in catastrophic consequences in applications, which makes the deep neural network hard to deploy in the \textit{open world}.
To address this issue, significant research efforts have been devoted to the
problem of OOD detection, which aims to establish a robust method for
identifying when the testing data is ``unknown.''

% Most effort is on exmaine the classifier weights
Specifically, OOD detection requires establishing a scoring function to
separate ID and OOD data. 
Designed for deep visual classifiers, the majority of works examine
activation patterns from the classification layer.
For example, a popular baseline is to leverage the maximum posterior
probability output of a softmax classifier to indicate ID-ness \cite{msp}, assuming that the network is more confident with its decision for ID data.
The idea is enhanced by neural network calibration techniques \cite{calib, odin, godin}, which aim to align network confidence with actual likelihood. The technique has been shown to improve OOD detection performance.
Other classifier-based scores explore unnormalized posterior probabilities
known as logits \cite{maxlogit, energyOOD}, and norms of gradients
backpropagated from the classification layer \cite{gradnorm, gradAlR}.

% It is important to combine both info, and why doing it in logit space is not a good idea. 
Recently, ViM \cite{vim} argues that both class-dependent and class-agnostic information can potentially facilitate the OOD detection task. 
Based on the idea, it designs a scoring function by mapping the
feature-space principle component residual to the logit space.
Despite the great performance achieved, ViM still requires supervision to
train the classification layer.
Meanwhile, advances in pretraining large-scale visual encoders using
contrastive learning techniques \cite{simCLR, supcon} have occured in recent
years.
These methods no longer rely on jointly training deep visual encoders with task heads, but instead formulate objectives directly in the feature space to produce high-quality visual representations.
%The common idea is to pull together the representations for similar concepts, while push apart that for unrelated data.
The learnt encoder can be applied to downstream tasks with little or no fine-tuning required~\cite{clip, cliport},
suggesting a new paradign for addressing computer vision problems.
While it is non-trivial to adopt classifier-based OOD detection methods directly to visual encoders, applying feature space methods is straightforward.
For example, SSD~\cite{ssd} applies Mahalanobis \cite{mahaOOD} distance in
the contrastive feature space, and demonstrates superior performance
compared to a typical classification visual encoder.

% What we do.
In this work, we aim to reason about both class-specific and class-agnostic information \textit{solely within the feature space}. 
To achieve this, we utilize Whitened Linear Discriminant Analysis (WLDA) to project visual features into two subspaces: a discriminative subspace and a residual subspace. 
The former contains compact class-discriminative signals, while the latter constrains shared information. 
We refer to these subspaces as \gSpace (\gSpaceS) and \hSpace (\hSpaceS), respectively.
Given the compactness of the  \gSpaceS space, we detect anomalies by measuring the distance to the nearest class center. 
Conversely, in the \hSpaceS space where in-distribution (ID) classes are entangled, we examine the distance to the centroid of all training data as an OOD indicator. 
The final proposed scoring function unifies the information from both subspaces by computing a weighted sum of the scores in each space.

% Our residual space is different than previous. Discovery: WDR space is very effective
While subspace techniques for OOD detection have been explored previously ~\cite{NuSA}, our approach differs significantly. Assuming ID data lies on a low-dimensional manifold, existing approaches measure the residual magnitude as an indicator for OOD-ness~\cite{subspace}.
In contrast, we assume the residual space captures rich class-shared information.
Furthermore, while past literature examines the residual to principles~\cite{vim} or classifier weights~\cite{NuSA},
we explore the remaining information to discriminative components.
This design enables us to jointly reason with both discriminative and residual information without relying on task heads, which is more applicable to stand-alone visual encoders.

% Summarize performance.
As shown in Fig. \ref{fig:teaser}, \methName achieves superior performance on large-scale ImageNet-1k benchmark compared to a wide range of baselines,
under both classic CNN and recent Visual Transformer (ViT) architectures.
In addition, our method surpasses other feature-space approaches in distinguishing novel concepts from stand-alone contrastive encoders, involving Supervised Contrastive (SupCon) model~\cite{supcon} and Contrastive Language-Image Pre-Training (CLIP) model \cite{clip}.
What's more, we discover that the Whitened Discriminative Residual Space is
more effective in identifying anomalous responses compared to other subspace techniques or even a subset of classifier-based scoring methods,
verifying the importance of the class-agnostic feature component for the task.

In summary, our contribution involves:
\begin{itemize}[leftmargin=*,noitemsep,nolistsep]
  \item A new OOD detection score, based on \WLDAS, that jointly considers
    class-discriminative and class-agnostic information solely within the
    feature space.  
  \item A new insight into the effectiveness of the \hSpace space, which
    captures the shared information stripped of discriminative signals, for
    detecting anomalies in OOD samples.
  \item New state-of-the-art results achieved by our method on the
    large-scale ImageNet OOD detection benchmark, for various visual
    classifiers (CNN and ViT) as well as contrastive visual encoders (SupCon
    and CLIP).
\end{itemize}

%------------------------------------------------------------------------
% Literature
\begin{figure*}[t!]
   \vspace{-0pt}
  % \vspace*{-0.0in}
  \centering
  \scalebox{1}{
    \begin{tikzpicture}
     \node[anchor=north west](graph) at (0in,0in)
      % {{\includegraphics[width=0.5\textwidth,trim={0.0in
      % 8.0in 4.0in 0.0in},clip=true]{figs/method.pdf}}};
      {\includegraphics[width=0.75\textwidth,trim={0.05in
      1.9in 0.05in 0.0in},clip=true]{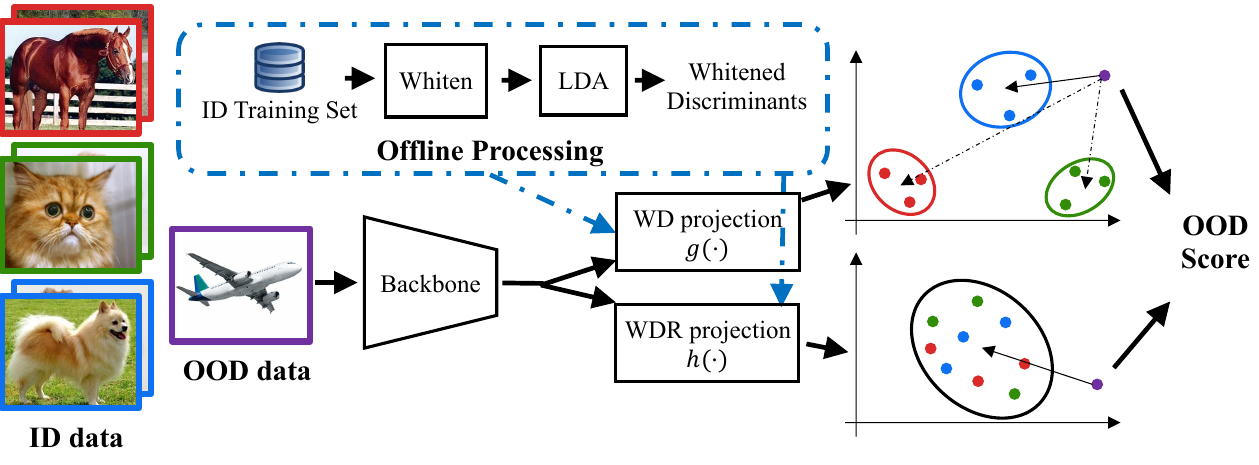}};
    % space annotation
    %  \node[anchor=south] at ($ (graph.north) + (0.3, -0.5)$)  {\small \bf \gSpaceS};
    %  \node[anchor=south] at ($ (graph.south) + (0.3, 0.0)$)  {\small \bf \hSpaceS};
%    \node[anchor=south] at ($ (graph.west) + (2.2, 1.3)$)  {\small $\enModel_\mParam(\hat{\btau{}}_{i-1})$};
%    \node[anchor=south] at ($ (graph.west) + (4.2, 1.3)$)  {\small $\enModel_\mParam(\hat{\btau{}}_{i})$};

    \end{tikzpicture}
  }
  \vspace{-6pt}
  \caption{\small
  \textbf{Overview of \methName method.}
  We detect an OOD sample by projecting the feature into \textit{\gSpace
  (\textbf{\gSpaceS})} and \textit{\hSpace (\textbf{\hSpaceS})}, where
  ID classes are maximally separated and closely clustered, respectively. 
  The projection functions are obtained from offline Whitened Linear
  Discriminant Analysis on the ID dataset.
  OOD samples are presumed to be far from the class clusters in \gSpaceS space 
  and the entire dataset clutter in \hSpaceS space. 
  Therefore, we formulate the OOD score as a combination of the distance
  to the nearest class centroid in the \gSpaceS space and to
  the entire dataset centroid in the \hSpaceS space.
  }
 \label{fig:method}
 \vspace{-10pt}
\end{figure*}

% fill to 2.5 page. 

\section{Related Work}

% Scoring functions for existing networks
% Network redesign/training constraints - include data
% Representation learning 

\paragraph{Scoring Functions for Pretrained Models}
One line of work explores scoring functions to distinguish inliers and
outliers, which is fundamental to the OOD detection task.  
Multiple designs have been proposed for pretrained deep visual classifiers  \cite{openMax, deepEnsembles,rottmann2020prediction,grcic2022densehybrid,  msp, energyOOD, maxlogit, gradnorm, gradAlR, react, vim, subspace, mahaOOD, knnOOD}.
The most straightforward design is Maximum Softmax Probability (MSP) score, which considers the network confidence as an uncertainty measurement.
ODIN \cite{odin} improves MSP's performance via two calibration techniques - input preprocessing and temperature scaling. 
Logit-space methods involve max logit \cite{maxlogit} and energy score \cite{energyOOD}. 
The latter is further enhanced by feature rectification \cite{energyOOD}.
Huang et al. \cite{gradnorm} and Lee et al. \cite{gradAlR} investigate gradient space, which demonstrates effectiveness in revealing ID and OOD distinction.
Those methods achieve great performance on several OOD detection benchmarks, but are tailored to the classification task and not applicable to pretrained visual encoders since they are dependent on classifier outputs.
On the other hand, methods based on feature space analysis, such as  Mahalanobis \cite{mahaOOD} and KNN \cite{knnOOD}, not only exhibit exceptional OOD detection performance for classification models, but also showcase applicability to stand-alone visual encoders, including SupCon \cite{supcon, ssd} and CLIP \cite{clip, limitOOD}.
Subspace analysis \cite{vim, subspace, NuSA} measures the residual information from low-dimensional manifold \cite{subspace} or column space of classifier weights. 
However, to achieve state-of-the-art performance, the result needs to be converted to classifier output space to factor in class-dependent information \cite{vim}.
In this work, we combine the residual and discriminative information solely
in the feature space via Linear Discriminant Analysis, which makes our
method applicable to any visual encorder.

\paragraph{Model Modifications \& Training Constraints}
Another branch of OOD detection methods trains the network to respond differently to ID and OOD data, by either modifying the network structure \cite{godin} or adding specialized training regularization \cite{MOS, oneDim, MCD}.
Along these lines, the most direct approach is to encourage the network to
give distinguishable predictions for outlier data, such as a uniform posterior probability \cite{outlierExposure}, lower energy \cite{energyOOD}, or confidence estimation from dedicated branch \cite{oodConf}.
to do so requires an auxiliary OOD training set, commonly known as Outlier Exposure (OE) \cite{outlierExposure}.
Several methods \cite{agnostophobia, outlierExposure} assume the availability of unknown data from outlier datasets, 
which can hardly cover all potention distribution shifts in the open world, and is not always feasible.
Lee et al. \cite{ganOE} utilizes generative models, such as Generative Adversarial Networks (GAN) \cite{gan} for outlier data synthesis.
However, generating high fidelity imagery induces optimization difficulties.
To avoid these obstacles, VOS \cite{vos} synthesize outliers in the feature
space, which can adapt to the ID feature geometry during training.
Although effective, these methods require network re-training to endow the
model with the ability to reject OOD data, which can be expensive and
intractable. Especially for models trained on large-scale datasets. 
On the other hand, our OOD scoring function design can be directly applied
on any pre-trained visual model.

%------------------------------------------------------------------------
% Methodology
\section{Methods}

% first define the goal of OOD problem - score function
The core of OOD detection lies in creating a scalar function $\score$ that assigns distinguishable scores to ID and OOD data. 
This allows the query data to be classified as either ID or OOD by applying a threshold to the score.
In this paper, we aim to assign higher scores for ID data.
Our approach involves disentangling class-discriminative and class-general
information from the features of a deep network's penultimate layer using \WLDA. 
We then jointly reason with both types of information to effectively identify OOD data. 
We first summarize the LDA method in Sec.\ref{sec:LDA}, then explain the
proposed \methName score based on LDA in Sec. \ref{sec:score} .

% First introduce Linear Discriminant Analysis (LDA) 
\subsection{Multiclass Linear Discriminant Analysis}
\label{sec:LDA}

The objective of Linear Discriminant Analysis (LDA) \cite{LDAtwo, LDAMulti} is to find a set of projection directions, termed \textit{discriminants}, along which multi-class data are maximally separated.
Specifically, given a training dataset $\{(\feat_i \in \Real^{\Dim}, \cls_i)\}_i^{\trainNum} $ of $\Dim$-dimensional features from $\clsNum$ classes ($\cls_i \in \{1, 2, \cdots, \clsNum\}$), the objective of LDA is to find the direction $\discriminant$ such that the \textit{Fisher discriminant criterion} \cite{LDA_ND}, defined as the ratio of inter-class variance over intra-class variance, is maximized:
\begin{equation}\label{eq:LDA}
	\max_{\discriminant} J(\discriminant) = 
      \frac{\discriminant^T \Sb \discriminant}
           {\discriminant^T \Sw \discriminant},
\end{equation}
where $\Sb$ and $\Sw$ are between-class scatter matrix and within-class scatter matrix. 
Denote the cardinality for class $c$ as $\trainNum_\cls$, then the
scatter matrices are formulated as:
\begin{align}\label{eq:Sw}
	\Sw &= \sum_{i=1}^\trainNum (\feat_i - \cen_{\cls_i})(\feat_i -
    \cen_{\cls_i})^T \ \text{and}\ \\
\label{eq:Sb}
	\Sb &= \sum_{c=1}^\clsNum \trainNum_c (\cen_c - \cen)(\cen_c - \cen)^T,
\end{align}
where $\cen$ is the center of the entire training dataset.

The optimization program defined in Eq. \ref{eq:LDA} is resolved by solving the generalized eigenvalue problem:
\begin{equation}\label{eq:GEig}
	\Sb \discriminant = \lambda \Sw \discriminant,
\end{equation}
where the generalized eigenvalue is equal to the Fisher criterion for the corresponding eigenvector.
Intuitively, a larger Fisher criterion suggests that ID classes are better separated.
Therefore, projecting original features along top discriminants, known as
the \textit{Foley-Sammon Transform (FST)}, maps the original feature to a low-dimensional subspace where ID classes are compactly clustered.
As described in the following section, we utilize FST to disentangle discriminative and residual information from visual feature space.

% Then describe our method
\subsection{\methName: Whitened Linear Discriminant Analysis for OOD detection}
\label{sec:score}

For a given extracted image feature $\bm{z}$,
Fig. \ref{fig:method} depicts the proposed \methName score to involve three 
steps:
(1) offline data whitening with the statistics derived from the training dataset;
(2) discriminative and residual space mapping with discriminants and
discriminant orthogonal complement set, which are estimated based on
offline LDA;
(3) a weighted sum of distances to ID cluster(s) in both spaces as the final \methName score.

\paragraph{Data Whitening}
We begin by whitening the feature prior to conducting LDA.
The application of whitened LDA has been extensively studied and implemented across various domains \cite{WLDA_face, WLDA_decor}.
According to Hariharan et al. \cite{WLDA_decor}, data whitening eliminates the correlation between feature elements and enhances the feature's capability to encode data similarity.
Furthermore, it improves the numerical conditioning of LDA, particularly
in scenarios with small sample sets.
In this paper, whitening is applied to the within-class covariance matrix of Eq. \ref{eq:Sw}. 
Suppose the eigenvalue decomposition for the covariance matrix on the feature space: $\bm{S}_{\bm{z}, w} = \eigVecMat_w \eigValMat_w \eigVecMat_w^T$, where $\eigVecMat_w \in \Real^{\Dim \times r}$, $\eigValMat_w \in \Real^{r \times r}$, $r$ denotes matrix rank, then the feature is whitened by:
\begin{equation}
	\feat = \bm{S}_{\bm{z}, w}^{-1/2} \bm{z} = \eigVecMat_w \eigValMat_w^{-1/2} \eigVecMat_w^T \bm{z}
\end{equation}

\paragraph{Discriminative and Residual Space Mapping}
To isolate the class-specific and class-agnostic components of the whitened 
visual feature, we leverage LDA to project the data onto two distinct subspaces.
The first subspace, referred to as the \textit{\gSpace (\gSpaceS)}, encodes the discriminative information of the data by compactly grouping samples from the same class and separating different classes.
The second subspace, referred to as the \textit{\hSpace (\hSpaceS)}, 
captures the residual of the discriminative information by clustering
together samples from all classes.

Following \cite{deepLDA}, we solve LDA by adding a multiple of identity
matrix to the within-class scatter matrix $\Sw + \rho \eye$. 
It stablizes small eigenvalues and ensures a sufficiently
well-conditioned scatter matrix.  It also converts the generalized
eigenvalue problem in Eq. \ref{eq:GEig} to an eigenvalue problem:
\begin{equation}
	(\Sw + \rho \eye)^{-1} \Sb \discriminant = \lambda \discriminant
\end{equation}
With discriminants solved from LDA, project the features into \gSpaceS
via Foley-Sammon Transformation.
To be specific, construct the projection matrix $\gMat =
[\discriminant_1, \discriminant_2, \cdots, \discriminant_{\discNum}] \in
\Real^{\Dim \times \discNum}$ by stacking the top $\discNum$
discriminants corresponding to the largest Fisher criterion.
The \gSpaceS projection is defined to be
\begin{equation}
	g(\feat) = \gMat^T \feat.
\end{equation}

To capture the class-agnostic information, 
project the data onto the subspace spanned by the orthogonal complements
of the top-$\discNum$ discriminants, e.g., to the \hSpaceS space.
The orthogonal complements correspond to projection directions with
lower Fisher criterion values, which implies that the separation between
classes is less significant compared to intra-class variance.
Thus, \hSpaceS space captures shared information among the ID classes,
which can be used to formulate additional constraints for ID data. 
Formally, suppose the eigendecomposition: $\gMat \gMat^T = \bm{Q}
\eigValMat_{\gMat} \bm{Q}^T$, then the \hSpaceS projection of a query feature 
is
\begin{equation}\label{eq:resScore}
	h(\feat) = (\eye - \bm{Q}\bm{Q}^T) \feat.
\end{equation}

\paragraph{\methName Score}
The \methName score combines ID data constraints in both \gSpaceS and
\hSpaceS spaces.  In the discriminative subspace, where discrepency
between known classes is maximized, all ID data is in close proximity to
some class cluster.
Formulate the \gSpaceS space OOD detection score as the distance
to the nearest class center $\cen_c^{\gSpaceS}$:
\begin{equation}\label{eq:gScore}
	s_g (\feat) = - \min_c \norm{g(\feat) - \cen_c^{\gSpaceS}}_2
\end{equation}
In the \hSpaceS space where inter-class discrepency is minimized,
the data from ID classes tends to scatter around a shared centroid.
Measure the OOD score in \hSpaceS space as the distance to the center of
all ID training data $\cen^{\hSpaceS}$:
\begin{equation}\label{eq:hScore}
	s_h(\feat) = - \norm{h(\feat) - \cen^{\hSpaceS}}_2
\end{equation}
The design is distinct from residual norm score from prior work~\cite{vim, subspace}, which assumes minimal information left from ID data that is irrelevant for the classification task.
Instead, we assume that the residual space contains abundant shared information, motivating us to formulate the distance-based score in Eq. \ref{eq:hScore}.

To incorporate information from both spaces, the \methName score
is defined as the weighted sum of both scores:
\begin{equation} \label{eq:finalScore}
	s(\feat) = s_g (\feat) + \hWeight s_h(\feat)
\end{equation}

%------------------------------------------------------------------------
% Experiments
%\input{tables/vit_results.tex}
%\input{tables/res50_results.tex}

\section{Experiments}
\label{sec:Exp}

This section compares the described approach with state-of-the-art
methods on a large-scale OOD detection benchmark to demonstrate the
effectiveness of the core concept. 
Evaluation is done for visual classifiers and stand-alone constrative
visual encoders.
Additionally, a comprehensive ablation study provides further insights
into \methName.

\paragraph{ID and OOD datasets}
Following recent work \cite{MOS}, testing involves a large-scale OOD detection
task with ImageNet-1k \cite{imagenet} as the ID dataset.
Six test OOD datasets are applied for evaluation, including:
\textit{SUN} \cite{SUN},
\textit{Places} \cite{Places},
\textit{iNaturalist} \cite{iNat},
\textit{Textures} \cite{Textures},
\textit{ImageNet-O} \cite{ImageNetO},
and \textit{OpenImage-O} \cite{vim}.
The first three (\textit{SUN}, \textit{Places}, \textit{iNaturalist}) use the subset curated by \cite{MOS} with non-overlapping categories \textit{w.r.t} the ID dataset.
Note that this evaluation is more comprehensive than previous OOD
detection literature investigating ImageNet benchmark \cite{vim,
knnOOD}, in the sense that they only adopt a subset of the above OOD datasets.
Utilizing a variety of OOD data sources encompasses a wider range of
distributional shift patterns, thus enabling a more comprehensive
evaluation of OOD detection methods.

\paragraph{Evaluation Metrics}
Evaluation uses two commonly adopted metrics that quantify a scoring
function's ability to distinguish ID and OOD data.
The first is \textit{Area under the Receiver Operating Characteristic
Curve (AUROC)}, a threshold-free metric that measures the area under the
plot of the true positive rate (TPR) against the false positive rate
(FPR) under varying classification thresholds. 
The AUROC metric is advantageous as it is invariant to the ratio of positive sample number to that of negative sample, making it suitable for evaluating OOD detection task, where the number of ID and OOD samples is imbalanced.
Higher value indicates better performance.
The second is \textit{False positive rate at $95\%$ true positive rate
(FPR95)} (smaller is better).

\paragraph{Models Settings and Hyperparameters}
Testing is done with classification models for ImageNet-1k \cite{imagenet} having various backbones. 
The first is a ResNet-50 backbone \cite{resnet}, the most widely applied convolutional neural network.
The second is a Vision Transformer (ViT), a transformer-based vision model that processes an input image as a sequence of patches. 
Following \cite{knnOOD}, we adopt the officially released ViT-B/16 architecture pretrained on ImageNet-21k and finetuned for classification on ImageNet-1k.

For stand-alone visual encoders, we test with Supervised Contrastive
(SupCon) \cite{supcon} and Contrastive Language-Image Pre-Training
(CLIP) \cite{clip} models.
The former is optimized to encourage smilarity between the embeddings of
samples from the same class, while maximizing the distance between them
for different classes. 
The latter is a multimodality representation learning method, trained to
pull together the features for matched image-text pairs, and push them away for non-matching pairs.
Both encoders adopt a ResNet-50 backbone with officially released weights.
Different from \cite{limitOOD}, we discard the language encoder from CLIP and perform OOD detection only in the visual feature space.
This removes the assumption of the availability of a textural ID class name or description, which is not always feasible in real-world application.
While both models utilize a projection head to a low-dimensional embedding space to formulate the training objectives, we leverage the backbone penultimate layer feature for better OOD detection performance following \cite{knnOOD}.
%For models with explicitly trained inner product between visual features, including SupCon and ViT (attention module), we L2-normalize the feature following \cite{ssd}. 
%We observe that it improves the performance for both \methName and Mahalanobis.

Per ViM \cite{vim}, we adopt different hyparameter settings according to feature dimension. 
For ResNet-50 encoder with $\Dim = 2048$ dimensional features, we set the number of discriminants as $\discNum=1000$ and the score weight in Eq. \ref{eq:finalScore} as $\hWeight=5$.
On the other hand, we adopt  $\discNum=500$ and $\hWeight=1$ for ViT feature space of $\Dim=768$ dimensionality.
When performing Linear Discriminant Analysis, we sample $\trainNum=200,000$ training images that are evenly distributed among all ID classes for the estimation of statisticcal quantities, such as means and scatter matrices.

\paragraph{Baseline Methods}
Comparison uses nine baselines that derive scores from pretrained models
without requiring network modification or finetuning.
Seven logit/probability-space methods are included: 
\textit{MSP} \cite{msp},
\textit{Energy} \cite{energyOOD},
\textit{ODIN} \cite{odin},
\textit{MaxLogit} \cite{maxlogit},
\textit{KLMatch} \cite{maxlogit},
\textit{ReAct} \cite{react},
and \textit{ViM} \cite{vim}.
For ReAct, we use Energy+ReAct with truncation percentile $p=99$. 
Two feature-space baselines are included:
\textit{Mahalanobis} \cite{mahaOOD} and
\textit{KNN} \cite{knnOOD}.
For the Mahalanobis method, we follow SSD \cite{ssd} to directly apply
the scoring function on the final layer feature without input-precossing
nor a multi-layer feature ensemble technique.

% Just put resnet50 and vit table together
\newcolumntype{g}{>{\columncolor{Gray}}c}
\begin{table*}
    \begin{subtable}[c]{\textwidth}
        \centering
        \setlength\tabcolsep{2. pt}
%        \begin{tabular}{@{}l@{}llc@{}lc@{}lc@{}lc@{}lc@{}lc@{}lc@{}lc@{}}
%	    \begin{tabular}{@{} l @{} lc@{} lc@{} lc@{} lc@{} lc@{} lc@{} lc@{}}
	    \begin{tabular}{@{}l @{} l c@{}c c@{}c c@{}c c@{}c c@{}c c@{}c| c@{}c}
            \toprule
%            \multirow{2}{*}{\begin{tabular}[c]{@{}c@{}}\textbf{Model}\end{tabular}}
             & \multirow{3}{*}{\begin{tabular}[c]{@{}c@{}}\textbf{Method}\end{tabular}} 
%             & \multirow{2}{*}{\begin{tabular}[c]{@{}c@{}}\textbf{Source}\end{tabular}} 
             & \multicolumn{2}{c}{\textbf{Textures}} 
             & \multicolumn{2}{c}{\textbf{SUN}} 
             & \multicolumn{2}{c}{\textbf{Places}} 
             & \multicolumn{2}{c}{\textbf{iNaturalist}} 
             & \multicolumn{2}{c}{\textbf{ImgNet-O}} 
             & \multicolumn{2}{c|}{\textbf{OpenImg-O}} 
             & \multicolumn{2}{c}{\textbf{Average}}                                                                                                                                                        \\
             &  & {\footnotesize FPR95$\downarrow$} & {\footnotesize AUROC$\uparrow$}
             	& {\footnotesize FPR95$\downarrow$} & {\footnotesize AUROC$\uparrow$}
             	& {\footnotesize FPR95$\downarrow$} & {\footnotesize AUROC$\uparrow$}
             	& {\footnotesize FPR95$\downarrow$} & {\footnotesize AUROC$\uparrow$}
             	& {\footnotesize FPR95$\downarrow$} & {\footnotesize AUROC$\uparrow$}
             	& {\footnotesize FPR95$\downarrow$} & {\footnotesize AUROC$\uparrow$}
             	& {\footnotesize FPR95$\downarrow$} & {\footnotesize AUROC$\uparrow$}
             	\\
            \cmidrule(r){1-2} \cmidrule(lr){3-14} \cmidrule{15-16}
%            \multirow{9}{*}{\begin{tabular}[c]{@{}c@{}}BiT\end{tabular}}
	    	 & \multicolumn{15}{c}{\textbf{Classifier-dependent methods}}  \\
             & {MSP}~\cite{msp} 
%             		& prob 
             		& $72.98$ & $74.92$
             		& $70.98$ & $78.75$ 
             		& $\underline{73.43}$ & $76.65$
             		& $60.90$ & $84.40$ 
             		& $95.65$ & $53.13$   
             		& $69.73$ & $81.17$
             		& $73.94$ & $74.84$
             		\\
             & {Energy}~\cite{energyOOD} 
%             		& logit 
             		& $95.74$ &  $48.60$
             		& $97.93$ &  $50.12$
             		& $97.77$ &  $48.90$
             		& $98.12$ &  $50.86$
             		& $92.80$ &  $48.23$
             		& $95.41$ &  $52.33$
             		& $96.30$ &  $49.84$
             		\\
             & {ODIN}~\cite{odin} 
%             		& prob+grad 
             		& $75.94$ &  $69.33$
             		& $75.51$ &  $74.05$
             		& $77.54$ &  $71.28$
             		& $68.60$ &  $79.88$
             		& $94.95$ &  $51.19$
             		& $73.98$ &  $76.15$
             		& $77.75$ &  $70.31$
             		\\
             & {MaxLogit}~\cite{maxlogit}     
%             		& logit 
             		& $75.92$ & $69.33$  
             		& $75.51$ & $74.05$
             		& $77.55$ & $71.28$
             		& $68.57$ & $79.88$
             		& $94.95$ & $51.19$
             		& $73.97$ & $76.15$
             		& $77.74$ & $70.31$
             		\\
             & {KLMatch}~\cite{maxlogit}  
%             		& prob  
             		& $57.57$ & $86.09$
             		& $70.36$ & $\underline{82.91}$
             		& $74.04$ & $\underline{80.65}$
             		& $46.83$ & $90.81$
             		& $89.75$ & $68.86$
             		& $58.21$ & $88.31$
             		& $66.13$ & $82.94$
             		\\
             & {ReAct}~\cite{react} 
%             		& feat+logit 
             		& $98.05$ & $34.51$
             		& $99.66$ & $23.68$
             		& $99.80$ & $22.86$
             		& $100.00$ & $23.13$
             		& $99.40$ & $37.31$
             		& $99.86$ & $23.86$
             		& $99.46$ & $27.56$
             		\\
             & {ViM \cite{vim}} 
%             		& feat+logit 
             		& $\underline{25.18}$ & $\underline{92.63}$
             		& $\underline{69.22}$ & $81.39$
             		& $74.90$ & $76.40$
             		& $\underline{30.02}$ & $\underline{93.38}$
             		& $\underline{76.15}$ & $\underline{77.08}$ 
             		& $\underline{46.70}$ & $\underline{88.60}$
             		& $\underline{53.70}$ & $\underline{84.91}$
             		\\
             \cmidrule(r){1-2} \cmidrule(lr){3-14} \cmidrule{15-16}
             & \multicolumn{15}{c}{\textbf{Feature space methods}}  \\
%             & {Residual \cite{vim}} 
%%             		& feat
%             	 	& $28.10$ & $92.04$
%             		& $72.49$ & $80.35$
%             		& $77.89$ & $75.67$
%             		& $34.05$ & $92.45$
%             		& $76.40$ & $76.29$
%             		& $51.41$ & $87.24$
%             		& $56.72$ & $84.01$
%             		\\
             & {Maha}~\cite{mahaOOD} 
%             		& feat+label
             		& $31.17$ & $91.62$ 
             		& $\underline{66.29}$ & $\underline{84.31}$ 
             		& $\underline{70.27}$ & $\underline{81.45}$
             		& $\underline{25.64}$ & $\underline{95.38}$
             		& $\underline{81.45}$ & $\underline{75.65}$ 
             		& $\bm{44.36}$ & $\bm{91.41}$ 
             		& $\underline{53.20}$ & $\underline{86.64}$
             		\\
             & {KNN~\cite{knnOOD}} 
%             		& feat
             	 	& $\bm{23.26}$ & $\bm{93.11}$
             	 	& $88.59$ & $74.01$
             	 	& $89.00$ & $71.07$
             	 	& $74.60$ & $85.83$
             	 	& $\bm{71.05}$ & $\bm{81.15}$ 
             	 	& $70.29$ & $84.01$
             	 	& $69.47$ & $81.53$
              	 	\\
             & {\textbf{\methName}}
             		& $\underline{29.20}$ & $\underline{91.90}$
             		& $\bm{56.83}$ & $\bm{86.74}$
             		& $\bm{64.40}$ & $\bm{83.13}$
             		& $\bm{22.39}$ & $\bm{95.59}$ 
             		& $81.60$ & $75.52$
             		& $\underline{44.67}$ &	$\underline{90.51}$
             		& \multicolumn{2}{l}{\cellcolor{lightgray}$\bm{49.85}$ $\bm{87.23}$}
             		\\
%            \cmidrule(r){1-3}\cmidrule(lr){4-11}\cmidrule(l){12-13}
%            \multirow{9}{*}{\begin{tabular}[c]{@{}c@{}}ViT\end{tabular}} \\
            \bottomrule
        \end{tabular}
        \caption{
            \textbf{ResNet-50~\cite{resnet}}. 
        }\label{tab:res50Clf}
    \end{subtable}
    \vspace{5pt}
	
    \begin{subtable}[c]{\textwidth}
        \centering
        \setlength\tabcolsep{2. pt}
%        \begin{tabular}{@{}l@{}llc@{}lc@{}lc@{}lc@{}lc@{}lc@{}lc@{}lc@{}}
%	    \begin{tabular}{@{} l @{} lc@{} lc@{} lc@{} lc@{} lc@{} lc@{} lc@{}}
	    \begin{tabular}{@{}l @{} l c@{}c c@{}c c@{}c c@{}c c@{}c c@{}c| c@{}c}
            \toprule
%            \multirow{2}{*}{\begin{tabular}[c]{@{}c@{}}\textbf{Model}\end{tabular}}
             & \multirow{3}{*}{\begin{tabular}[c]{@{}c@{}}\textbf{Method}\end{tabular}} 
%             & \multirow{2}{*}{\begin{tabular}[c]{@{}c@{}}\textbf{Source}\end{tabular}} 
             & \multicolumn{2}{c}{\textbf{Textures}} 
             & \multicolumn{2}{c}{\textbf{SUN}} 
             & \multicolumn{2}{c}{\textbf{Places}} 
             & \multicolumn{2}{c}{\textbf{iNaturalist}} 
             & \multicolumn{2}{c}{\textbf{ImgNet-O}} 
             & \multicolumn{2}{c|}{\textbf{OpenImg-O}} 
             & \multicolumn{2}{c}{\textbf{Average}}                                                                                                                                                        \\
             &  & {\footnotesize FPR95$\downarrow$} & {\footnotesize AUROC$\uparrow$}
             	& {\footnotesize FPR95$\downarrow$} & {\footnotesize AUROC$\uparrow$}
             	& {\footnotesize FPR95$\downarrow$} & {\footnotesize AUROC$\uparrow$}
             	& {\footnotesize FPR95$\downarrow$} & {\footnotesize AUROC$\uparrow$}
             	& {\footnotesize FPR95$\downarrow$} & {\footnotesize AUROC$\uparrow$}
             	& {\footnotesize FPR95$\downarrow$} & {\footnotesize AUROC$\uparrow$}
             	& {\footnotesize FPR95$\downarrow$} & {\footnotesize AUROC$\uparrow$}
             	\\
            \cmidrule(r){1-2} \cmidrule(lr){3-14} \cmidrule{15-16}
%            \multirow{9}{*}{\begin{tabular}[c]{@{}c@{}}BiT\end{tabular}}
	    	 & \multicolumn{15}{c}{\textbf{Classifier-dependent methods}}  \\
             & {MSP}~\cite{msp} 
%             		& prob 
             		& $52.43$ & $85.42$
             		& $53.22$ & $86.93$ 
             		& $57.75$ & $85.72$
             		& $13.66$ & $97.00$ 
             		& $51.75$ & $85.81$   
             		& $31.99$ & $92.48$
             		& $43.47$ & $88.89$
             		\\
             & {Energy}~\cite{energyOOD} 
%             		& logit 
             		& $36.13$ & $91.25$
             		& $\underline{34.44}$ & $\underline{93.28}$
             		& $\bm{42.80}$ & $\bm{90.98}$
             		& $5.60$ &  $98.94$
             		& $\underline{30.30}$ & $93.36$
             		& $16.06$ & $96.87$
             		& $27.56$ & $94.11$
             		\\
             & {ODIN}~\cite{odin} 
%             		& prob+grad 
             		& $38.57$ &  $90.86$
             		& $37.45$ &  $92.81$
             		& $44.68$ &  $\underline{90.66}$
             		& $6.03$ &   $98.81$
             		& $33.50$ &  $92.69$
             		& $17.83$ &  $96.54$
             		& $29.68$ &  $93.73$
             		\\
             & {MaxLogit}~\cite{maxlogit}     
%             		& logit 
             		& $38.56$ & $90.86$
             		& $37.45$ & $92.81$
             		& $44.68$ & $\underline{90.66}$
             		& $6.03$ &  $98.81$
             		& $33.50$ & $92.69$
             		& $17.83$ & $96.54$
             		& $29.68$ & $93.73$
             		\\
             & {KLMatch}~\cite{maxlogit}  
%             		& prob  
             		& $51.22$ & $85.12$
             		& $56.04$ & $85.45$
             		& $61.08$ & $83.86$
             		& $13.68$ & $96.32$
             		& $49.90$ & $85.62$
             		& $31.38$ & $91.93$
             		& $43.88$ & $88.05$
             		\\
             & {ReAct}~\cite{react} 
%             		& feat+logit 
             		& $\bm{36.35}$ &  $91.17$
             		& $34.55$ &  $93.22$
             		& $\underline{43.32}$ &  $\underline{90.83}$
             		& $5.61$ &   $98.94$
             		& $\underline{30.30}$ &  $93.40$
             		& $\underline{16.01}$ &  $96.88$
             		& $27.69$ &  $94.07$
             		\\
             & {ViM \cite{vim}} 
%             		& feat+logit 
             		& $38.67$ & $\underline{91.38}$
             		& $\bm{32.47}$ & $\bm{93.41}$
             		& $44.23$ & $89.86$
             		& $\underline{1.40}$ & $\underline{99.68}$
             		& $31.80$ & $\underline{94.05}$ 
             		& $16.61$ & $\underline{97.10}$
             		& $\underline{27.53}$ & $\underline{94.25}$
             		\\
             \cmidrule(r){1-2} \cmidrule(lr){3-14} \cmidrule{15-16}
             & \multicolumn{15}{c}{\textbf{Feature space methods}}  \\
%             & {Residual \cite{vim}} 
%%             		& feat
%             	 	& $49.06$ & $89.35$
%             		& $38.63$ & $91.29$
%             		& $52.81$ & $86.56$
%             		& $1.51$ & $99.62$
%             		& $40.10$ & $92.21$
%             		& $25.41$ & $95.58$
%             		& $34.59$ & $92.44$
%             		\\
             & {Maha}~\cite{mahaOOD} 
%             		& feat+label
             		& $\underline{36.61}$ & $\underline{91.67}$ 
             		& $35.37$ & $92.89$ 
             		& $46.08$ & $89.55$
             		& $\underline{0.96}$ & $\underline{99.78}$
             		& $30.45$ & $\underline{94.22}$ 
             		& $\bm{13.85}$ & $\bm{97.50}$ 
             		& $\underline{27.22}$ & $\underline{94.27}$
             		\\
             & {KNN~\cite{knnOOD}} 
%             		& feat
             	 	& $38.28$ & $90.74$
             	 	& $46.08$ & $90.73$
             	 	& $54.50$ & $87.54$
             	 	& $6.75$ & $98.70$
             	 	& $38.95$ & $92.53$ 
             	 	& $20.59$ & $96.12$
             	 	& $34.19$ & $92.72$
              	 	\\
             & {\textbf{\methName}}
             		& $\underline{36.58}$ & $\bm{91.79}$
             		& $\underline{32.62}$ & $\underline{93.34}$
             		& $\underline{43.74}$ & $89.91$
             		& $\bm{0.89}$ & $\bm{99.81}$ 
             		& $\bm{30.15}$ & $\bm{94.36}$
             		& $\underline{14.30}$ & $\underline{97.44}$
             		& \multicolumn{2}{l}{\cellcolor{lightgray}$\bm{26.38}$ $\bm{94.44}$} 
             		\\
%            \cmidrule(r){1-3}\cmidrule(lr){4-11}\cmidrule(l){12-13}
%            \multirow{9}{*}{\begin{tabular}[c]{@{}c@{}}ViT\end{tabular}} \\
            \bottomrule
        \end{tabular}
        \caption{
            \textbf{ViT-B/16~\cite{ViT}}. 
        }\label{tab:vitClf}
    \end{subtable}
	\vspace{-5pt}
    \caption{
    \textbf{Results on ResNet-50~\cite{resnet} and ViT~\cite{ViT} classification models}. 
    We test all methods on six OOD datasets and compute the average performance.
   Both metrics AUROC and FPR95 are in percentage.
   We highlight the best performance in bold, and underline the 2nd and 3rd ones.
   Our method consistently outperforms all classifier-dependent or feature-space baselines under both network architectures in terms of average performance.
   }\label{tab:clfResults}
   \vspace{-5pt}
\end{table*}

\subsection{OOD Detection Performance on Classifiers}
The first experiment presents the OOD detection comparison with
classification models based on ResNet-50 and ViT backbones.
Tab. \ref{tab:res50Clf} and Tab. \ref{tab:vitClf} provide the quantitative 
results, consisting of outcomes for each OOD dataset and average
performance across them.
The best AUROC and FPR95 are in bold, with second and third rank underlined.
On average across all OOD datasets, \methName exhibits superior
performance compared to the baselines for both ViT and ResNet-50 models.
This validates the efficacy of our OOD score function design.

\paragraph{Consistency across OOD distributions.}
Tajwar et al \cite{noSOTA} shows on small-scale OOD benchmarks that
existing OOD detectors do not have a consistent performance across OOD
data sources.
Consistency is important as the OOD data distribution is unpredictable in the 
open world.
\methName consistently achieves top performance across all OOD datasets.
As an illustration, it achieves top-3 performance in 11 out of 12
metrics (2 - AUROC and FPR95 - for each of the 6 OOD datasets) when used
with the ViT-B/16 model, with 6 of them being the highest compared to
the baseline methods. 
Close baselines are ViM and Maha (both achieve $7$ top-3 including $2$ top-1). 
but their top-1 results are achieved for one dataset, whereas \methName
does so for three.  Similar outcomes occur for ResNet-50. 
Examining both discriminative and residual information promotes the
detection of a broader range of OOD patterns.

\paragraph{Robustness across encoder architectures.}
Another property of \methName is its model-agnostic nature. 
Its consistently high performance across network types demonstrates
an ability to handle the diverse feature manifolds induced by distinct
network architectures.

\subsection{Results on Contrastive Visual Encoders}
\label{sec:expEncResults}

The \methName implementation is applied to the SupCon \cite{supcon}
and CLIP \cite{clip} models to verify its applicability on modern visual
encoders.
Since classifier-dependent baselines are not applicable to the
stand-alone visual feature extractors, baseline comparison involves only
the feature-space baselines - Mahalanobis \cite{mahaOOD} and KNN \cite{knnOOD}.
Tab. \ref{tab:encoders} collects the results.
\methName surpasses the baselines for both models,
suggesting that it is more effective in identifying OOD patterns across
the feature spaces induced by diverse representation learning objectives.

% DUPLICATED PARAGRAPH.
%In order to demonstrate the applicability of our method on modern visual encoders, we evaluate it on SupCon \cite{supcon} and CLIP \cite{clip} models. 
%Since classifier-dependent baselines are not suitable for stand-alone visual feature extractors, 
%we only compare with two feature-space baselines - Mahalanobis \cite{mahaOOD} and KNN \cite{knnOOD}. 
%The results, shown in Table \ref{tab:encoders}, demonstrate that \methName outperforms both baselines on both models. 
%This suggests that our method is effective in identifying OOD patterns from the feature space trained with diverse representation objectives.

\subsection{Understanding \methName}
\label{sec:expUnderstand}

\paragraph{Discriminant Residual v.s. Principle Residual}
To confirm the efficacy of \hSpaceS space for OOD detection, it is 
compared to other subspace residual designs.
ViM \cite{vim} bases their method on the residual of principle space,
with empirical evidence on its effectiveness over the classifier null space
method \cite{NuSA}. Comparative evaluation of this approach against the
residual score in Eq. \ref{eq:hScore} on classification models is
provided in Tab. \ref{tab:subspaceCompare}.

The \hSpaceS space scoring function outperforms the previous subspace
technique in terms of AUROC and FPR95 for both ResNet-50 and ViT
architectures.
Additionally, \textit{without} the discriminative information, 
it performs competitively compared to SOTA in Tab. \ref{tab:clfResults}.
In particular, the \hSpaceS space score achieves the third highest AUROC
on ResNet-50 classifier, behind the integrated \methName and Mahalanobis
scores.  This placing supports the claim that \hSpaceS space is highly
informative for OOD detection.

\begin{table}
    \centering
    \begin{tabular}{|l|cc|cc|}
%        \toprule
		\hline
        \multirow{2}{*}{\begin{tabular}[c]{@{}c@{}}\textbf{Method}\end{tabular}} 
        & \multicolumn{2}{c|}{\textbf{SupCon}~\cite{supcon}} 
        & \multicolumn{2}{c|}{\textbf{CLIP}~\cite{clip}}  
        \\
         & {\small FPR95$\downarrow$} & {\small AUROC$\uparrow$}
        & {\small FPR95$\downarrow$} & {\small AUROC$\uparrow$}                    
         \\
%        \midrule
		\hline
        Mahalanobis  & $46.95$  & $89.78$  & $78.00$   & $75.31$
        \\
        KNN  & $42.51$  & $90.35$   & $82.59$   & $67.22$
         \\
        \textbf{\methName}  & $\bm{40.10}$ & $\bm{90.89}$   & $\bm{77.57}$   & $\bm{75.74}$ 
         \\
%        \bottomrule
		\hline
    \end{tabular}
    \caption{
        \textbf{Results on SupCon \cite{supcon} and CLIP \cite{clip} visual encoders.}
        Our method surpasses other feature-space methods on these representation learning models.
        This table reports only average performance, while detailed results for each dataset can be found in the Supplementary material.
    }\label{tab:encoders}
    \vspace{-3pt}
\end{table}
\begin{table}
    \centering
    \begin{tabular}{|l|cc|cc|}
%        \toprule
		\hline
        \multirow{2}{*}{\begin{tabular}[c]{@{}c@{}}\textbf{Method}\end{tabular}} 
        & \multicolumn{2}{c|}{\textbf{ResNet-50}} 
        & \multicolumn{2}{c|}{\textbf{ViT}}  
        \\
         & {\small FPR95$\downarrow$} & {\small AUROC$\uparrow$}
        & {\small FPR95$\downarrow$} & {\small AUROC$\uparrow$}                    
         \\
%        \midrule
		\hline
        PR  & $56.72$  & $84.01$ & $34.59$   & $92.44$ 
        \\
        \textbf{\hSpaceS}  & $\bm{53.74}$ & $\bm{86.56}$   & $\bm{30.35}$   & $\bm{93.72}$ 
         \\
%        \bottomrule
		\hline
    \end{tabular}
    \caption{
		\textbf{Comparison between subspace techniques.}        
	    Our proposed \hSpaceS space is more effective than the Principle Residual (PR) space \cite{vim} for the OOD detection task.
	    Furthermore, its AUROC on ResNet-50 outperforms 8 out of 9 baselines, which demonstrates that \hSpaceS is informative for OOD detection.
    }\label{tab:subspaceCompare}
 	\vspace{-8pt}
\end{table}

\begin{table}[t!]
  \vspace*{-0.0in}
  \centering
  
  \setlength\tabcolsep{3.pt} 
  \renewcommand{\arraystretch}{1.}
  \begin{threeparttable}
  \begin{tabular}{|c c | c c | c c |}
  		\hline
	\multicolumn{2}{|c|}{Config} 
%	Config  &
	& \multicolumn{2}{|c|}{ResNet-50} 
	& \multicolumn{2}{|c|}{ViT} 
		\\
 		\hline
 	Whiten$^\dagger$  & Dist
 	& {\small FPR95 $\downarrow$} &  {\small AUROC $\uparrow$}
 	& {\small FPR95 $\downarrow$} & {\small AUROC $\uparrow$}
		\\
% 		\hhline{|======|}
		\hline
	\xmark  & Maha & 53.65 & 86.20 & 29.81 & 93.47 
		\\ 
	\xmark  & Eucl & 74.56 & 81.17 & 32.21 & 93.52 
	    \\
	 \cmark & Maha & 49.85 & 87.23 & 26.60 & 94.40
	    \\
	   \hline\hline
	 \cmark & Eucl & 49.86 & 87.23 & 26.49 & 94.41
	 \\
		\hline
  \end{tabular}
  \end{threeparttable}
  \caption{ 
  \textbf{Ablation on Data whitening}.
  $\dagger$: The option of data whitening before DLA.
  Whitening the data prior to DLA improves over no whitening (\xmark + Eucl) or whitening after DLA (\xmark + Maha).
  Furthermore, with prior data whitening, Euclidean (\cmark+Eucl) performs similar as Mahalanobis (\cmark+Maha) while requiring less computation,
  which justifies our design. 
  }
  \label{tab:ablateWhiten}
  \vspace{-8pt}
\end{table}

\paragraph{Effect of Feature Whitening}
We evaluate \methName with or without feature whitening on the ResNet-50 and 
ViT classification models.
To isolate the effect of feature decorrelation on LDA, we also ablate on
Euclidean distance vs Mahalanobis distance (whitening + Euclidean) in 
Eq. \ref{eq:gScore} and Eq. \ref{eq:hScore}.
Tab. \ref{tab:ablateWhiten} collects the results, which indicate
that feature whitening improves (FR95) performance by a large margin.
Replacing Euclidean distance with Mahalanobis after LDA slightly improves the results without whitening, but is insufficent to fill the gap.
This suggests that feature whitening facilitates isolation of
class-specific and residual information for LDA, which is crucial for
our method.  Mahalanobis \textit{after} \WLDAS has similar performance
as Euclidean, but requires additional scatter matrix estimates in
subspaces.  Therefore our design is justified.

\begin{figure}[t!]
   \vspace{-0pt}
  \centering
  \scalebox{0.8}{
    \begin{tikzpicture}
     \node[anchor=north west](graph) at (0in,0in)
      {{\includegraphics[width=0.5\textwidth,trim={0.0in
      0.0in 0.0in 0.0in},clip=true]{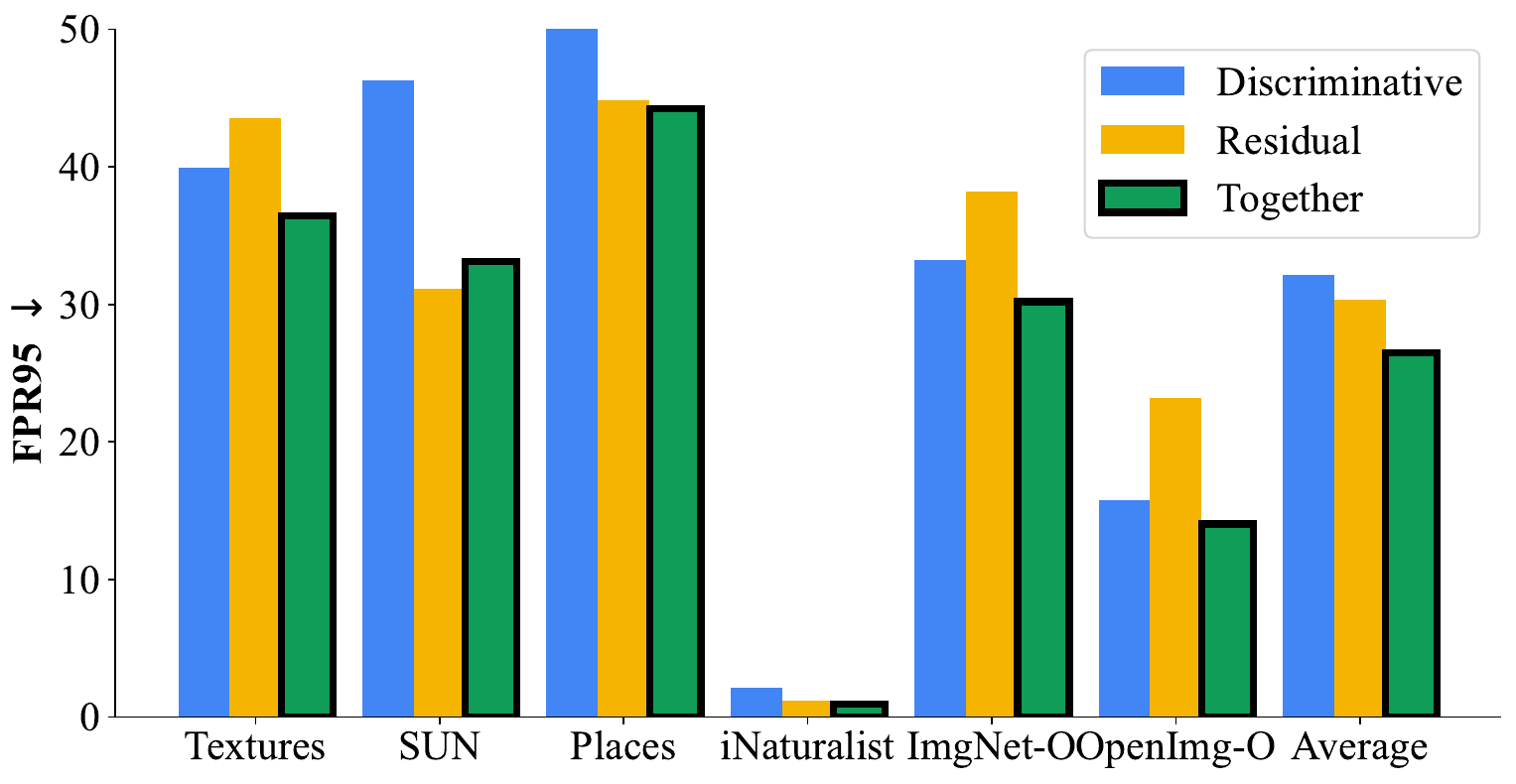}}};
    % space annotation
%     \node[anchor=south] at ($ (graph.north) + (0.3, -0.5)$)  {\small \bf \gSpaceS};

    \end{tikzpicture}
  }
  \vspace{-10pt}
  \caption{\small
  \textbf{The individual performance in \gSpaceS and \hSpaceS space.}
  Our results demonstrate that two subspaces are effective in detecting OOD data from distinct distributions. 
  Moreover, the integrated score (black border) yields a superior performance compared to each individual,
  evidencing that our method unifies class-discriminative and class-agnostic information.
  }
 \label{fig:spaceBreak}
 \vspace{-5pt}
\end{figure}

\paragraph{Contribution by \gSpaceS and \hSpaceS Spaces}
To understand the role played by each subspace, 
we break down the scoring function and evaluate the OOD detection \textit{solely} within \gSpaceS or \hSpaceS,
 with Eq. \ref{eq:gScore} or Eq. \ref{eq:hScore} as OOD scores, separately.
The FPR95 performance is illustrated in Fig. \ref{fig:spaceBreak}.
Our results demonstrate that two subspaces are effective in detecting different OOD patterns. 
Specifically, \gSpaceS outperforms \hSpaceS on three OOD datasets (SUN, Places, INaturalist), while underperforming on the others (Textures, ImageNet-O, OpenImage-O). 
Additionally, the integrated score achieves better performance than each individual space on average, 
indicating that our method effectively leverages the complementary information provided by both subspaces.
Unlike ViM \cite{vim}, we combine the class-specific and class-agnostic
information for OOD detection without relying on task heads, which
improves its general use.

%% Remove since the efficacy of hSpace is already discussed in residual comparison paragraph.
%Another interesting discovery is \textit{the efficacy of \hSpace for OOD detection}.
%The achieved average results are better than those of the discriminative subspace, 
%which is \red{unexpected} since the residual information is typically assumed to have less influence on the training classification objective.
%\red{
%It indicates that OOD data might trigger activations more in the residual space or null space, which can potentailly have a broader impact on OOD Generalization research community. \red{Some exist work could investigate this already, cite some if find any}
%}

\begin{figure}[t!]
   \vspace{-0pt}
  % \vspace*{-0.0in}
  \centering
  \scalebox{0.8}{
    \begin{tikzpicture}
     \node[anchor=north west](graph) at (0in,0in)
      {{\includegraphics[width=0.5\textwidth,trim={0.0in
      0.0in 0.0in 0.0in},clip=true]{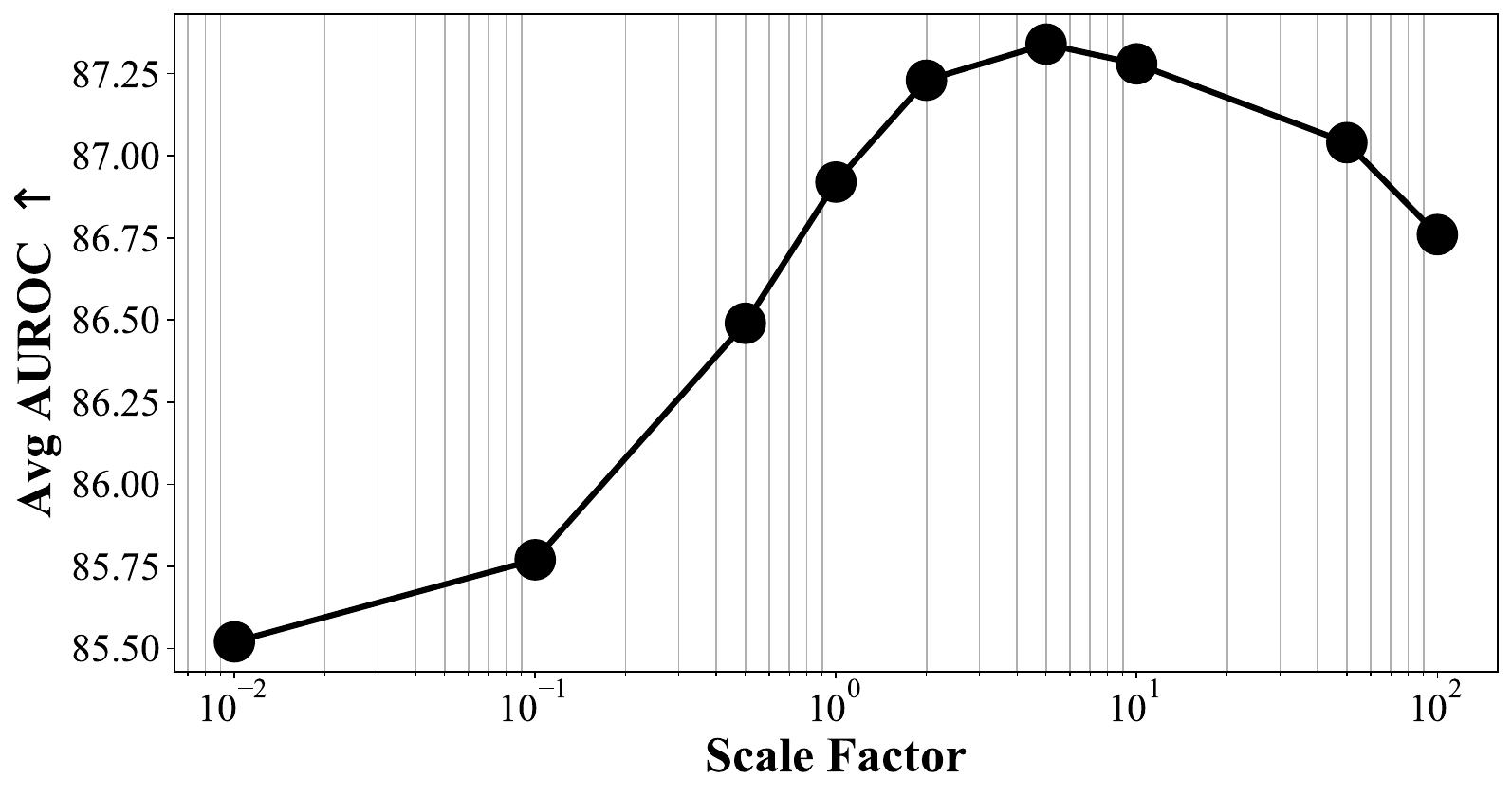}}};
    % space annotation
%     \node[anchor=south] at ($ (graph.north) + (0.3, -0.5)$)  {\small \bf \gSpaceS};

    \end{tikzpicture}
  }
  \vspace{-12pt}
  \caption{\small
  \textbf{Ablate on scoring scale factor}.
  The integrated performance is better than individual subspace over a wide range of scales, 
  indicating that our simple linear combination is effective in factoring in information from both subspaces for OOD detection.
  }
 \label{fig:ablateScale}
 \vspace{-8pt}
\end{figure}

\paragraph{Effect of scale factor $\hWeight$}
We ablate on the scale factor $\hWeight$ in Eq. \ref{eq:finalScore} on the ResNet-50 classification model.
We fix the number of discriminants $\discNum = 1000$ and test with different scales from the set: $\hWeight \in \{0.01, 0.1, 0.5, 1, 2, 5, 10, 100\}$.
Fig. \ref{fig:ablateScale} depicts the average AUROC under varied scaling factor.
When the scaling factor is extreme, the score function biases towards one subspace over the other, leading to lower performance at both ends of the curve. 
The integrated scoring function outperforms the individual subspace
scores across a wide range of scales, indicating that the proposed
linear combination is effective at considering information from both
subspaces.

\begin{figure}[t!]
%   \vspace{-10pt}
  % \vspace*{-0.0in}
  \centering
  \scalebox{0.8}{
    \begin{tikzpicture}
     \node[anchor=north west](graph) at (0in,0in)
      {{\includegraphics[width=0.5\textwidth,trim={0.0in
      0.0in 0.0in 0.0in},clip=true]{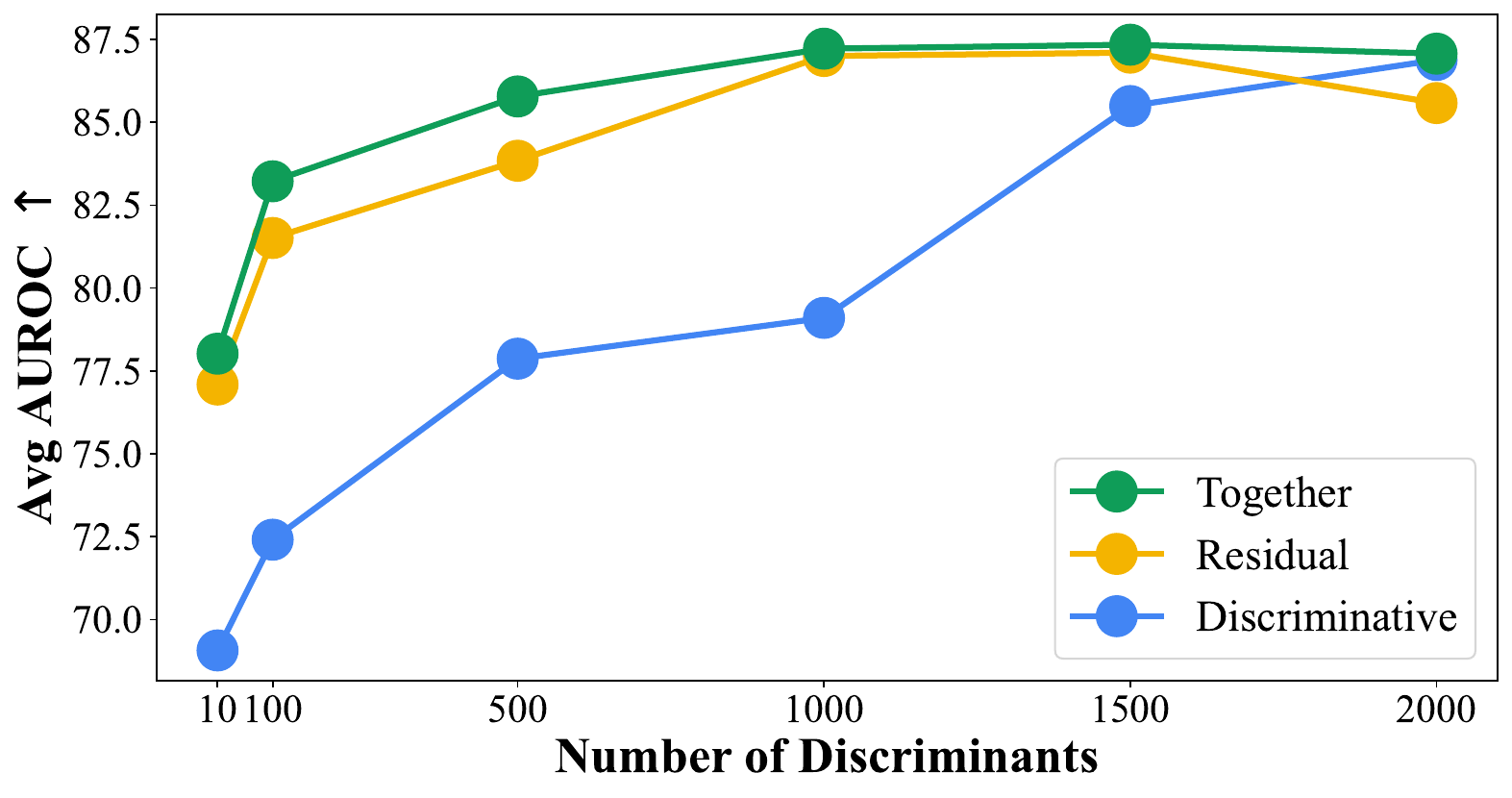}}};
    % space annotation
%     \node[anchor=south] at ($ (graph.north) + (0.3, -0.5)$)  {\small \bf \gSpaceS};

    \end{tikzpicture}
  }
  \vspace{-12pt}
  \caption{\small
  \textbf{Ablate on Discriminant Number $\discNum$.}
 Our method achieves consistently high performance when the discriminant number is sufficiently large, enabling complete disentanglement of discriminative and residual information.
  }
 \label{fig:ablateNumDisc}
 \vspace{-5pt}
\end{figure}

\paragraph{Effect of Discriminant Number $\discNum$}
We also ablate on the effect of discriminant number $\discNum$, by testing our method with $\discNum \in \{10, 100, 500, 1000, 1500, 2000\}$ on the same model.
For each discriminant number, we test with varied scaling factor within the same set as above, and report the highest AUROC results. 
We further report individual performance from each subspace.
The results are demonstrated in Fig. \ref{fig:ablateNumDisc}.
As the discriminant number increases, \textit{the individual performance in both \gSpaceS and \hSpaceS space improves as the result of better separation between discriminative and residual information}.
The trend stops when the number is sufficiently large, as the separation saturates.
As a result, the intergrated performance becomes stable with increased discriminant number.

\begin{figure}[t!]
   \vspace{-0pt}
  % \vspace*{-0.0in}
  \centering
  \scalebox{0.8}{
    \begin{tikzpicture}
     \node[anchor=north west](graph) at (0in,0in)
      {{\includegraphics[width=0.5\textwidth,trim={0.0in
      0.0in 0.0in 0.0in},clip=true]{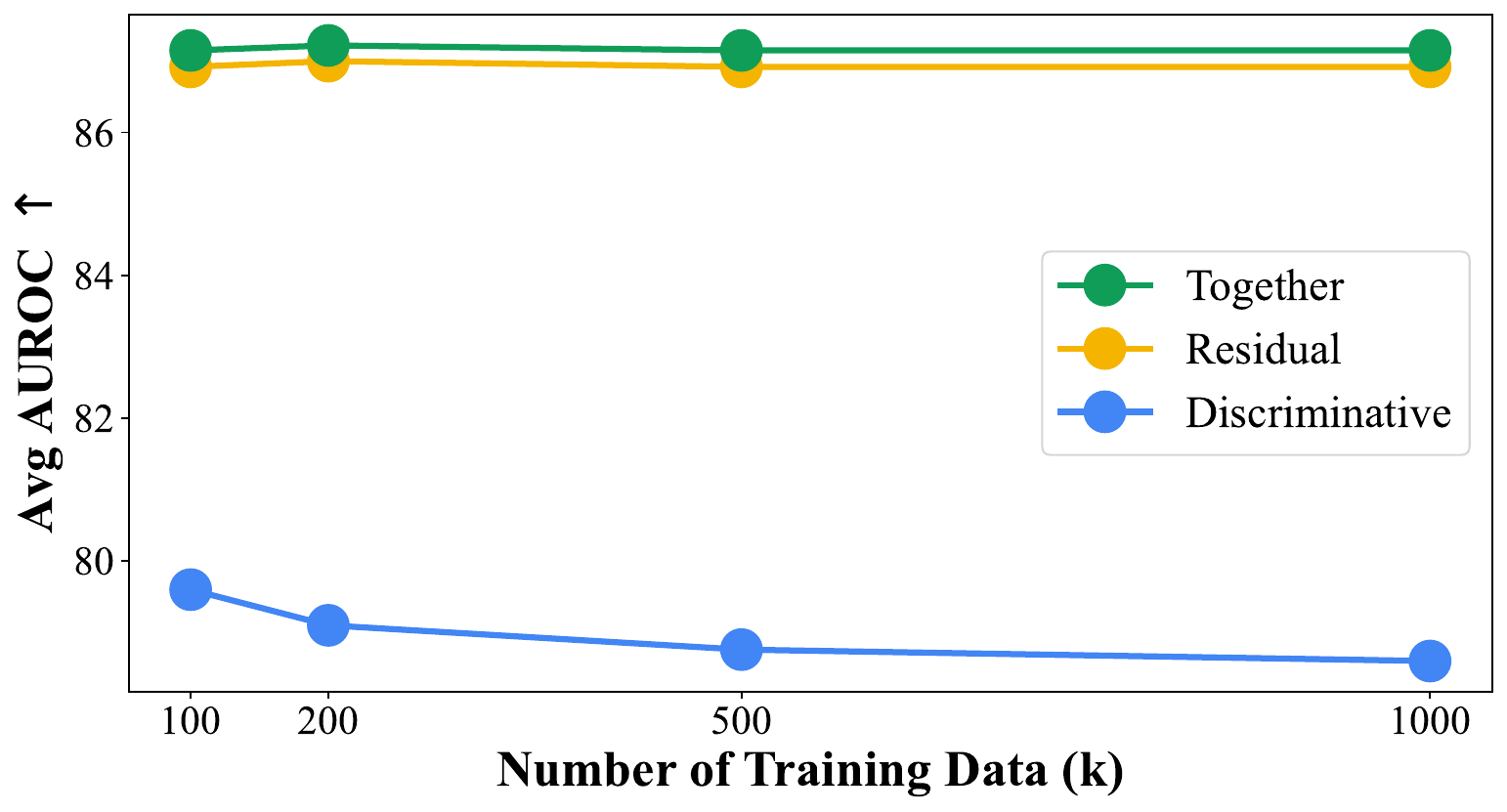}}};
    % space annotation
%     \node[anchor=south] at ($ (graph.north) + (0.3, -0.5)$)  {\small \bf \gSpaceS};

    \end{tikzpicture}
  }
  \vspace{-12pt}
  \caption{\small
  \textbf{Ablate on Training Data Number.}
  Our method achieves near-optimal performance using only $200K$ training data out of over $1000K$, 
  indicating that it is not excessively data-hungry.
  }
 \label{fig:ablateTrainNum}
 \vspace{-8pt}
\end{figure}

\paragraph{Robustness against Training Data Number $\trainNum$}
Here we explore the effects of the training data quantity $\trainNum$ on
the performance of \methName.
We vary the quantity and plot the resulting average AUROC for individual subspace performance and integrated performance in Fig. \ref{fig:ablateTrainNum}.
Near-optimal results occur with $20\%$ of the training data ($200K$ out of 
over $1000K$), indicating the data efficiency of our method.

%------------------------------------------------------------------------
% Conclusion
\section{Conclusion}

This paper presents a new OOD detection method based on \WLDA (\WLDAS),
named \methName.
It jointly reasons with class-specific and class-agnostic information by
disentangling the discriminative and residual information from the
feature space via \WLDAS.
Comprehensive evaluation shows that {\methName} establishes superior
results on multiple large-scale benchmark, demonstrating robustness
across model architectures as well as representation learning
objectives. 
Analysis reveals the efficacy of \hSpace (\hSpaceS) on OOD detection
compared to other subspace techniques, showing the importance of
understanding the behavior of OOD activation in the residual subspace.

\balance
{\small
\bibliographystyle{ieee_fullname}
\bibliography{reference}
}

%------------------------------------------------------------------------
% Supplementary
\clearpage
\appendix

%%%%%%%%%%%%%%%%%%%%%%%%%%%%%%%%%%%%%%%%%%%%%%%%%%%%%%%%%%%%%%%%%%%%%%%%%%%%%%%%%%%%%%%
\section{Model details}
For ResNet-50 and ViT-B/16 classifiers, we adopt the feature encoder trained with a single-layer classification head on the ImageNet-1k training dataset.
ViT-B/16 refers to the base model variant (layer=12, dimension $\Dim=768$, heads=12) with $16\times16$ input patch size.
For the CLIP visual encoder, we adopt the ResNet-50 model trained with ViT-B/32 language encoder. We discard the language model and only use visual encoder in our experiments. 
The input data is cropped and resized to $224\times224$ for ResNet models (including ResNet-50 classification encoder, SupCon, and CLIP), and $384 \times 384$ for ViT-B/16. 
For both Mahalanobis~\cite{mahaOOD} and {\methName}, we $L2$-normalize
the feature for models directly trained on inner products between visual features (including SupCon model with Supervised Contrastive loss on normalized feature, and ViT with attention mechanism). 
We found that a normalized feature space enhances OOD detection performance when the inner product between image features is trained to encode similarity.

%%%%%%%%%%%%%%%%%%%%%%%%%%%%%%%%%%%%%%%%%%%%%%%%%%%%%%%%%%%%%%%%%%%%%%%%%%%%%%%%%%%%%%%
\section{Baseline details}

\paragraph{Mahalanobis}
We remove the input preprocessing and feature ensemble techniques proposed in the original paper~\cite{mahaOOD} for small-scale benchmarks,
as we find that they compromise the performance on large-scale benchmarks.
Instead, we follow SSD~\cite{ssd} and apply the Mahalanobis distance
directly to the penultimate layer feature.
$200,000$ random training samples are used for calculating the precision matrix and class-wise centroids.

\paragraph{KNN}
For all models, we $L2$-normalize the features following the original work~\cite{knnOOD}.
The KNN score is calculated on $200,000$ random training data.
The nearest number is downscaled proportionally based on $k=1000$ for the full training set.

\paragraph{ReAct}
Following the practice of \cite{vim}, we use the most effective
Energy+ReAct setting.
We also adopt rectification percentile $p=99$ instead of $p=90$ from original work~\cite{react} for better performance.

\paragraph{Principle Residual (PR)}
The settings for Principle Residual (PR) baseline evaluated in Sec. \ref{sec:expUnderstand} are adopted from ViM~\cite{vim}.
Prior to principle component estimation, we center the features based on classification layer weights and bias.
$1000$ principle components are used when the feature dimension is
greater than $1500$ (ResNet-50, SupCon,CLIP), otherwise $512$ principle
components are used (for ViT).

\begin{figure}[h!]
    \vspace{-0pt}
   % \vspace*{-0.0in}
   \centering
   \scalebox{0.85}{
     \begin{tikzpicture}
      \node[anchor=north west](graph) at (0in,0in)
       {\includegraphics[width=0.5\textwidth,trim={0.05in
       0.0in 2.5in 0.0in},clip=true]{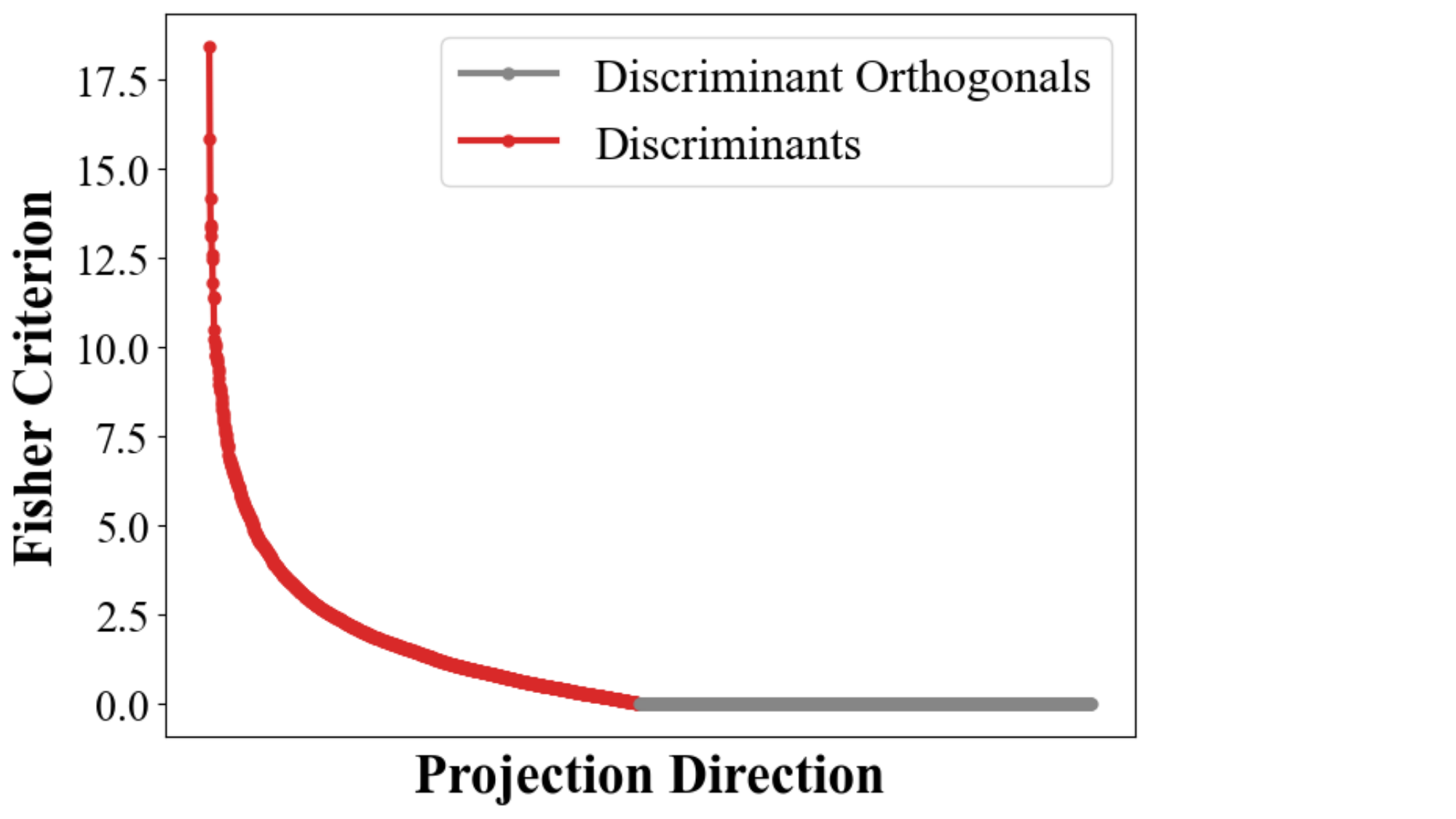}};
     % space annotation
     %  \node[anchor=south] at ($ (graph.north) + (0.3, -0.5)$)  {\small \bf \gSpaceS};
     %  \node[anchor=south] at ($ (graph.south) + (0.3, 0.0)$)  {\small \bf \hSpaceS};
 %    \node[anchor=south] at ($ (graph.west) + (2.2, 1.3)$)  {\small $\enModel_\mParam(\hat{\btau{}}_{i-1})$};
 %    \node[anchor=south] at ($ (graph.west) + (4.2, 1.3)$)  {\small $\enModel_\mParam(\hat{\btau{}}_{i})$};
 
     \end{tikzpicture}
   }
   \vspace{-8pt}
   \caption{\small
   \textbf{Empirical Fisher Criterion (FC) values for ResNet-50 classification model.}
     FC values for discriminants are non-trivial whereas that for discriminant orthogonals approach zero. 
     This verifies our assumption that WLDA disentangle discriminative and residual information from the feature space.
   }
  \label{fig_sup:empiricalFC}
  \vspace{-10pt}
 \end{figure}

\section{Empirical Fisher Criterion Value}

To empirically verify our assumption that WLDA disentangles discriminative and residual information, 
we compare the Fisher Criterion (FC) values for discriminants and
discriminant residuals from ResNet-50 classification model trained on
ImageNet; see Fig. \ref{fig_sup:empiricalFC}. 
The FC values for discriminants are non-trivial, indicating separation of ID
features along the directions.
On the other hand, the projections along discriminant orthogonal directions are non-separable, as the FC values in those subspaces are close to zero.
This verifies our assumption that WLDA separate class-specific and class-agnostic information, and explains the superior performance of \methName method.

%%%%%%%%%%%%%%%%%%%%%%%%%%%%%%%%%%%%%%%%%%%%%%%%%%%%%%%%%%%%%%%%%%%%%%%%%%%%%%%%%%%%%%%
\section{Detailed Results on SupCon and CLIP}
Sec. \ref{sec:expEncResults} provides the average results for
{\methName} and the feature-space baselines on SupCon and CLIP visual encoders.
Here, Tab. \ref{tab:encResultsSupp} gives the AUROC and FPR95 measures on
all six OOD datasets 

% Just put resnet50 and vit table together
\newcolumntype{g}{>{\columncolor{Gray}}c}
\begin{table*}[h!]
    \begin{subtable}[c]{\textwidth}
        \centering
        \setlength\tabcolsep{2. pt}
%        \begin{tabular}{@{}l@{}llc@{}lc@{}lc@{}lc@{}lc@{}lc@{}lc@{}lc@{}}
%	    \begin{tabular}{@{} l @{} lc@{} lc@{} lc@{} lc@{} lc@{} lc@{} lc@{}}
	    \begin{tabular}{@{}l @{} l c@{}c c@{}c c@{}c c@{}c c@{}c c@{}c| c@{}c}
            \toprule
%            \multirow{2}{*}{\begin{tabular}[c]{@{}c@{}}\textbf{Model}\end{tabular}}
             & \multirow{3}{*}{\begin{tabular}[c]{@{}c@{}}\textbf{Method}\end{tabular}} 
%             & \multirow{2}{*}{\begin{tabular}[c]{@{}c@{}}\textbf{Source}\end{tabular}} 
             & \multicolumn{2}{c}{\textbf{Textures}} 
             & \multicolumn{2}{c}{\textbf{SUN}} 
             & \multicolumn{2}{c}{\textbf{Places}} 
             & \multicolumn{2}{c}{\textbf{iNaturalist}} 
             & \multicolumn{2}{c}{\textbf{ImgNet-O}} 
             & \multicolumn{2}{c|}{\textbf{OpenImg-O}} 
             & \multicolumn{2}{c}{\textbf{Average}}                                                                                                                                                        \\
             &  & {\footnotesize FPR95$\downarrow$} & {\footnotesize AUROC$\uparrow$}
             	& {\footnotesize FPR95$\downarrow$} & {\footnotesize AUROC$\uparrow$}
             	& {\footnotesize FPR95$\downarrow$} & {\footnotesize AUROC$\uparrow$}
             	& {\footnotesize FPR95$\downarrow$} & {\footnotesize AUROC$\uparrow$}
             	& {\footnotesize FPR95$\downarrow$} & {\footnotesize AUROC$\uparrow$}
             	& {\footnotesize FPR95$\downarrow$} & {\footnotesize AUROC$\uparrow$}
             	& {\footnotesize FPR95$\downarrow$} & {\footnotesize AUROC$\uparrow$}
             	\\
            \cmidrule(r){1-2} \cmidrule(lr){3-14} \cmidrule{15-16}
%            \multirow{9}{*}{\begin{tabular}[c]{@{}c@{}}BiT\end{tabular}}
             & {Maha}~\cite{mahaOOD} 
%             		& feat+label
             		& $14.80$ & $95.62$ 
             		& $63.09$ & $86.76$ 
             		& $68.93$ & $84.20$ 
             		& $33.41$ & $95.06$ 
             		& $\bm{65.50}$ & $83.00$ 
             		& $35.96$ & $94.05$ 
             		& $46.95$ & $89.78$ 
             		\\
             & {KNN~\cite{knnOOD}} 
%             		& feat
             	 	& $15.18$ & $95.62$ 
             		& $47.97$ & $89.29$ 
             		& $58.33$ & $85.45$ 
             		& $30.30$ & $94.83$ 
             		& $66.10$ & $\bm{83.88}$ 
             		& $37.18$ & $93.05$ 
             		& $42.51$ & $90.35$ 
              	 	\\
             & {\textbf{\methName}}
             	 	& $\bm{13.94}$ & $\bm{95.84}$ 
             		& $\bm{47.87}$ & $\bm{89.40}$ 
             		& $\bm{58.21}$ & $\bm{86.34}$ 
             		& $\bm{21.49}$ & $\bm{96.21}$ 
             		& $66.50$ & $83.27$ 
             		& $\bm{32.59}$ & $\bm{94.26}$ 
             		& \multicolumn{2}{l}{\cellcolor{lightgray}$\bm{40.10}$ $\bm{90.89}$} 
             		\\
%            \cmidrule(r){1-3}\cmidrule(lr){4-11}\cmidrule(l){12-13}
%            \multirow{9}{*}{\begin{tabular}[c]{@{}c@{}}ViT\end{tabular}} \\
            \bottomrule
        \end{tabular}
        \caption{
            \textbf{SupCon~\cite{supcon}}. 
        }\label{tab:res50ClfSupp}
    \end{subtable}
    \vspace{5pt}
	
    \begin{subtable}[c]{\textwidth}
        \centering
        \setlength\tabcolsep{2. pt}
%        \begin{tabular}{@{}l@{}llc@{}lc@{}lc@{}lc@{}lc@{}lc@{}lc@{}lc@{}}
%	    \begin{tabular}{@{} l @{} lc@{} lc@{} lc@{} lc@{} lc@{} lc@{} lc@{}}
	    \begin{tabular}{@{}l @{} l c@{}c c@{}c c@{}c c@{}c c@{}c c@{}c| c@{}c}
            \toprule
%            \multirow{2}{*}{\begin{tabular}[c]{@{}c@{}}\textbf{Model}\end{tabular}}
             & \multirow{3}{*}{\begin{tabular}[c]{@{}c@{}}\textbf{Method}\end{tabular}} 
%             & \multirow{2}{*}{\begin{tabular}[c]{@{}c@{}}\textbf{Source}\end{tabular}} 
             & \multicolumn{2}{c}{\textbf{Textures}} 
             & \multicolumn{2}{c}{\textbf{SUN}} 
             & \multicolumn{2}{c}{\textbf{Places}} 
             & \multicolumn{2}{c}{\textbf{iNaturalist}} 
             & \multicolumn{2}{c}{\textbf{ImgNet-O}} 
             & \multicolumn{2}{c|}{\textbf{OpenImg-O}} 
             & \multicolumn{2}{c}{\textbf{Average}}                                                                                                                                                        \\
             &  & {\footnotesize FPR95$\downarrow$} & {\footnotesize AUROC$\uparrow$}
             	& {\footnotesize FPR95$\downarrow$} & {\footnotesize AUROC$\uparrow$}
             	& {\footnotesize FPR95$\downarrow$} & {\footnotesize AUROC$\uparrow$}
             	& {\footnotesize FPR95$\downarrow$} & {\footnotesize AUROC$\uparrow$}
             	& {\footnotesize FPR95$\downarrow$} & {\footnotesize AUROC$\uparrow$}
             	& {\footnotesize FPR95$\downarrow$} & {\footnotesize AUROC$\uparrow$}
             	& {\footnotesize FPR95$\downarrow$} & {\footnotesize AUROC$\uparrow$}
             	\\
            \cmidrule(r){1-2} \cmidrule(lr){3-14} \cmidrule{15-16}
%            \multirow{9}{*}{\begin{tabular}[c]{@{}c@{}}BiT\end{tabular}}
             & {Maha}~\cite{mahaOOD} 
%             		& feat+label
             		& $54.11$ & $89.77$ 
             		& $\bf{81.36}$ & $77.45$ 
             		& $83.87$ & $78.21$ 
             		& $97.74$ & $56.41$ 
             		& $76.50$ & $\bf{74.89}$ 
             		& $\bf{74.42}$ & $\bf{75.13}$ 
             		& $78.00$ & $75.31$ 
             		\\
             & {KNN~\cite{knnOOD}} 
%             		& feat
             		& $59.61$ & $88.92$ 
             		& $89.65$ & $69.86$ 
             		& $90.33$ & $70.76$ 
             		& $99.59$ & $36.52$ 
             		& $\bf{75.35}$ & $73.48$ 
             		& $80.98$ & $63.77$ 
             		& $82.59$ & $67.22$ 
              	 	\\
             & {\textbf{\methName}}
             		& $\bf{54.10}$ & $\bf{89.85}$ 
             		& $81.45$ & $\bf{78.33}$ 
             		& $\bf{81.54}$ & $\bf{80.14}$ 
             		& $\bf{96.81}$ & $\bf{57.69}$ 
             		& $76.95$ & $74.38$ 
             		& $74.59$ & $74.05$ 
             		& \multicolumn{2}{l}{\cellcolor{lightgray}$\bm{77.57}$ $\bm{75.74}$} 
             		\\
%            \cmidrule(r){1-3}\cmidrule(lr){4-11}\cmidrule(l){12-13}
%            \multirow{9}{*}{\begin{tabular}[c]{@{}c@{}}ViT\end{tabular}} \\
            \bottomrule
        \end{tabular}
        \caption{
            \textbf{CLIP~\cite{clip}}. 
        }\label{tab:vitClfSupp}
    \end{subtable}
	\vspace{-5pt}
    \caption{
    \textbf{Results on SupCon~\cite{supcon} and CLIP~\cite{clip} visual encoders}. 
    We test all methods on six OOD datasets and compute the average performance.
   Both metrics AUROC and FPR95 are in percentage.
   We highlight the best performance in bold.
   {\methName} more consistently outperforms the alternatives for both
   encoders in terms of average FPR95 and
   AUROC.\label{tab:encResultsSupp}}
\end{table*}

%%%%%%%%%%%%%%%%%%%%%%%%%%%%%%%%%%%%%%%%%%%%%%%%%%%%%%%%%%%%%%%%%%%%%%%%%%%%%%%%%%%%%%%
\section{OOD detection in the Embedding space}
Both SupCon and CLIP formulate the contrastive loss in a low-dimensional feature space obtained from a projection head.
As explained in Sec. \ref{sec:Exp}, we follow KNN~\cite{knnOOD} and SSD~\cite{mahaOOD} to apply all feature-space methods on the penultimate layer feature space for better performance.
To further verify the claim, we test all feature-space methods, including KNN, Mahalanobis, and the proposed \methName, in the embedding spaces.
For hyperparameters in \methName, we choose $\discNum=512$ and $\hWeight=1$ for CLIP embeddings with $\Dim=1024$ dimensions, and $\discNum=50$ and $\hWeight=1$ for SupCon model with embedding dimension as $\Dim=128$. 

\begin{table}[H]
    \centering
    \setlength\tabcolsep{3.5 pt}
    \begin{tabular}{|l|c|cc|cc|}
%        \toprule
		\hline
        \multirow{2}{*}{\begin{tabular}[c]{@{}c@{}}\textbf{Method}\end{tabular}} 
        & \multirow{2}{*}{\begin{tabular}[c]{@{}c@{}}\textbf{Space}\end{tabular}} 
        & \multicolumn{2}{c|}{\textbf{SupCon}~\cite{supcon}} 
        & \multicolumn{2}{c|}{\textbf{CLIP}~\cite{clip}}  
        \\
         & & {\small FPR95$\downarrow$} & {\small AUROC$\uparrow$}
        & {\small FPR95$\downarrow$} & {\small AUROC$\uparrow$}                    
         \\
%        \midrule
		\hline
        Maha  & \multirow{3}{*}{\begin{tabular}[c]{@{}c@{}}Embed\end{tabular}} 
        		& $54.74$  & $86.61$  & $97.06$   & $67.83$
        \\
        KNN  & & $55.96$  & $86.40$  & $96.42$   & $61.81$
         \\
        \textbf{WDisc}  & & $53.10$  & $87.22$  & $92.25$   & $63.26$
         \\
         \hline
        Maha  & \multirow{3}{*}{\begin{tabular}[c]{@{}c@{}}\makecell{Last \\ Layer}\end{tabular}} 
        	& $46.95$  & $89.78$  & $78.00$   & $75.31$
        \\
        KNN  & & $42.51$  & $90.35$   & $82.59$   & $67.22$
         \\
        \textbf{WDisc}  & & $\bm{40.10}$ & $\bm{90.89}$   & $\bm{77.57}$   & $\bm{75.74}$ 
        \\
%        \bottomrule
		\hline
		
    \end{tabular}
    \caption{
        \textbf{Comparison between penultimate feature space and embedding space for SupCon \cite{supcon} and CLIP \cite{clip} for all feature-space methods.}
      Low-dimensional embeddings are less information for OOD detection compared to penultimate layer features, suggesting potential loss of information critical for the task.
    }\label{tab:embedResults}
    \vspace{-10pt}
\end{table}

Comparision between performance in the embedding space and penultimate feature space is in Tab. \ref{tab:embedResults}, where all methods suffer from performance degradation in the embedding space.
The results show that distance between visual features in the embedding space do not imply similarity, which is potentially caused by lost information due to limited dimensionality.

\end{document}